\documentclass[11pt]{article}

\usepackage[final]{acl}

\usepackage{times}
\usepackage{latexsym}

\usepackage[T1]{fontenc}

\usepackage[utf8]{inputenc}

\usepackage{microtype}

\usepackage{inconsolata}

\usepackage{hyperref}
\usepackage{url}
\usepackage{extpfeil}
\usepackage{booktabs}
\usepackage{xcolor}
\usepackage[most]{tcolorbox}
\usepackage{inconsolata}
\usepackage{etoolbox}
\usepackage[table]{xcolor}
\usepackage{multirow}
\usepackage{graphicx}
\usepackage{caption}
\usepackage{subcaption}
\usepackage{wasysym}
\usepackage{marvosym}
\definecolor{ManBlue}{RGB}{222,240,255}
\newcommand{\bb}[1]{\cellcolor{ManBlue}\textbf{{#1}}}
\definecolor{PromptBack}{HTML}{FFF1B8}
\definecolor{PromptFrame}{HTML}{C9A100}
\definecolor{PromptTitle}{HTML}{000000}
\newcommand{\EqualCalls}{\mathrm{EqualCalls}}
\newcommand{\EC}{\EqualCalls(C_{\text{calls}},G_{\text{calls}})}
\newcommand{\equalcontrib}{\textsuperscript{*}}
\newcommand{\corrauth}{\textsuperscript{\Letter}}
\newcommand{\projlead}{\textsuperscript{\dag}}
\usepackage{enumitem}

\lstdefinestyle{promptlisting}{
  breaklines=true,
  breakatwhitespace=false,
  postbreak=\mbox{\textcolor{PromptFrame}{\(\hookrightarrow\)}\space},
  basicstyle=\ttfamily\scriptsize, 
  columns=fullflexible,
  keepspaces=true,
  tabsize=2,
  showstringspaces=false,
  xleftmargin=0pt,
  xrightmargin=0pt
}

\newtcolorbox{promptbox}[1][]{
  enhanced, breakable,
  colback=PromptBack,
  colframe=PromptFrame,
  coltitle=PromptTitle,
  fonttitle=\bfseries\footnotesize,
  fontupper=\footnotesize,             
  title=#1,
  boxrule=0.6pt,                       
  arc=1.5pt,
  left=4pt,right=4pt,top=4pt,bottom=4pt,  
  before upper={\setlength{\parindent}{0pt}\setlength{\parskip}{2pt}}
}
\title{Failure makes the agent stronger: Enhancing Accuracy \\ through Structured Reflection for Reliable Tool Interactions}

\author{%
  Junhao Su$^{1}$\equalcontrib\;
  Yuanliang Wan$^{1}$\equalcontrib\;
  Junwei Yang$^{1}$\equalcontrib\AND
  Hengyu Shi$^{2}$\;
  Tianyang Han$^{2}$\;
  Yurui Qiu$^{1}$\projlead\corrauth\;
  Junfeng Luo$^{1,2}$\corrauth
  \\[0.2cm]                                
  \textsuperscript{1}Vision Agent Team, Meituan~~~
  \textsuperscript{2}MeiGen AI Team, Meituan\\
  \texttt{\{qiuyurui,luojunfeng\}@meituan.com}\\
  \equalcontrib Equal Contrib ~~~
  \corrauth Corresponding Authors ~~~
  \projlead Project Leader \\
}

\begin{document}

\maketitle

\begin{abstract}
Tool-augmented large language models (LLMs) are typically trained via supervised imitation learning or coarse-grained reinforcement learning, approaches that primarily optimize one-shot tool calls. Existing practices of self-reflection largely rely on heuristic prompting or unidirectional reasoning traces: the model is encouraged to “think more,” rather than to treat error diagnosis and correction as a learnable capability. This makes them fragile in multi-turn interaction settings—once a call fails, the model tends to repeat the same mistake instead of recovering.
To address this issue, we propose structured reflection, which transforms the “from error to repair” process into a first-class, controllable, and trainable action. The agent produces a concise yet precise reflection process: specifically, the model diagnoses the error based on evidence from the previous step and then proposes a correct and executable follow-up call. During training, we combine DAPO and GSPO's objective functions and design a more principled reward mechanism tailored to tool calling, optimizing the stepwise strategy Reflect $\to$ Call $\to$ Final.
To evaluate this capability, we introduce Tool-Reflection-Bench, a lightweight benchmark dataset that programmatically verifies structural validity, executability, parameter correctness, and result consistency. Tasks in the benchmark are constructed as miniature trajectories of Erroneous Call $\to$ Reflection $\to$ Corrected Call and are split into disjoint training and testing sets.
Experiments on BFCL v3 and Tool-Reflection-Bench show that our method achieves significant improvements in multi-turn tool-call success rates and error recovery, while also reducing redundant calls. These results demonstrate that making reflection explicit and treating it as an optimization objective can substantially enhance the reliability of tool interaction, providing a reproducible pathway for agents to grow stronger by learning from failure. Our Code and Dataset is available at: \url{https://github.com/MeiGen-AI/Tool-Reflection-Bench}
\end{abstract}

\begin{figure*}[htbp]
  \centering
  \begin{subfigure}{0.495\textwidth}
    \centering
    \includegraphics[width=\linewidth]{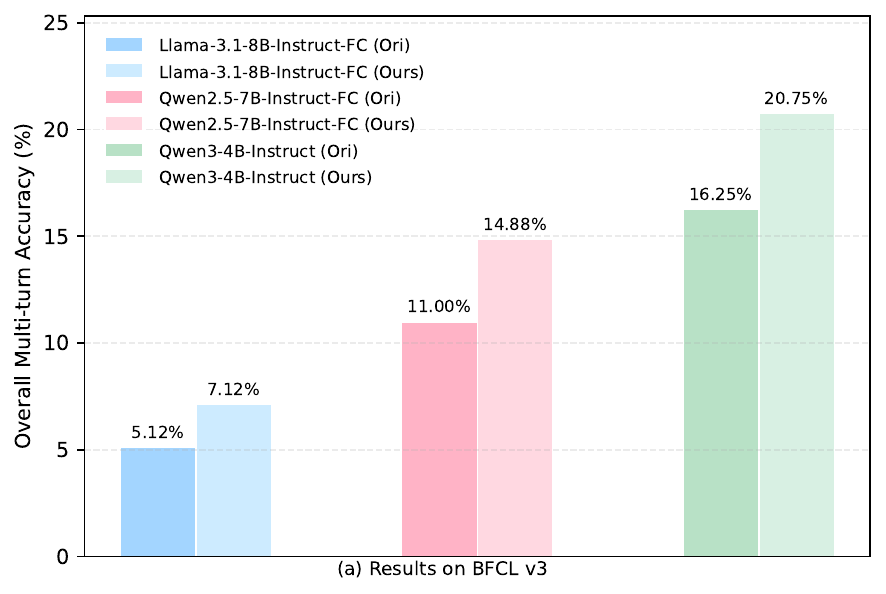}
    \caption{Results on BFCL v3}\label{fig:bfcl}
  \end{subfigure}\hfill
  \begin{subfigure}{0.495\textwidth}
    \centering
    \includegraphics[width=\linewidth]{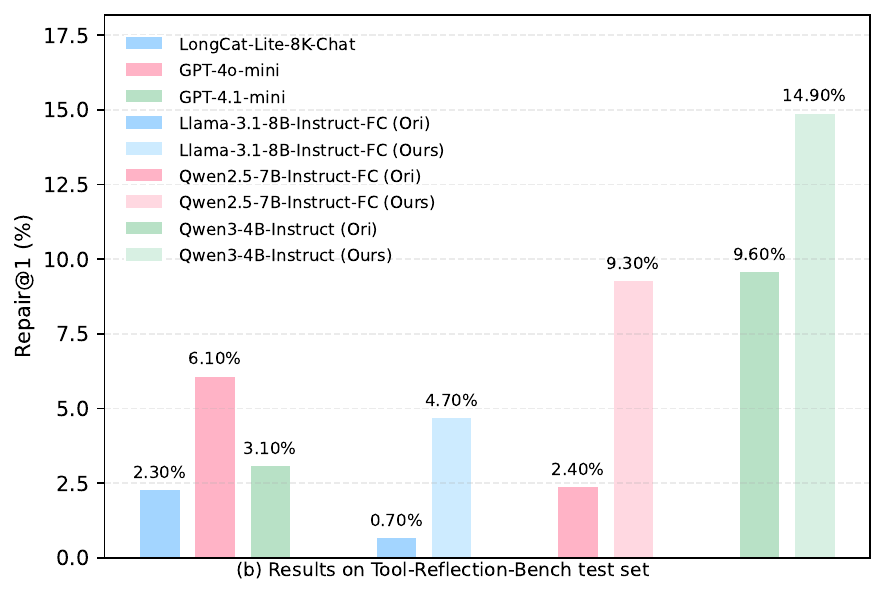}
    \caption{Results on Tool-Reflection-Bench test set}\label{fig:trbench}
  \end{subfigure}
  \caption{In the experiments on BFCL v3 and Tool-Reflection-Bench, our method significantly improves the multi-turn tool-calling accuracy of several open-source LLMs on BFCL v3.
At the same time, it substantially enhances the error-repair rate for tool calls on the Tool-Reflection-Bench test set, achieving performance that even surpasses that of closed-source LLMs with comparable parameter sizes.}
  \label{fig:two-bars}
\end{figure*}

\section{Introduction}
The integration of external tools with large language models through tool calling represents a significant breakthrough in the development of agents. It transforms large language models from mere text generators into highly practical tools for interacting with humans \cite{1,2}, significantly enhancing the ability of AI agents to solve complex real-world tasks \cite{3,toolbench,5}. Tool calling bridges the gap between the vast internal knowledge of LLMs and external resources, enabling LLMs to access up-to-date information, perform delicate computations, and more, thereby unlocking their broad potential for applications across multiple domains \cite{6,7,8}.

Currently, the training of tool-call capabilities in large language models typically relies on supervised fine-tuning and reinforcement learning \cite{9,10}, where these methods optimize the ability for single-turn tool calls through carefully designed reward mechanisms. However, these approaches face several challenges in the context of tool calling. First, the issue of rewards in tool calling is particularly prominent—small errors in parameter selection or formatting often render the entire function call invalid, thus limiting the effective learning signal \cite{11}. Second, existing methods generally rely on unidirectional reasoning, which, while sufficient for simpler scenarios, has clear limitations: when LLMs make mistakes during tool calls, they often struggle to locate the root cause of the error \cite{12}. While generating correct function calls is crucial, it is even more important for LLMs to learn how to identify and correct their own mistakes \cite{13}.

To address the above-mentioned issues, we propose an innovative reflection process aimed at error localization and correction through explicit reflection steps, which differs from existing forward reasoning methods. Specifically, we design a process in which the LLM intentionally makes mistakes during tool calls, carefully crafts reflection content based on the errors, and then generates the correct call. Through this approach, we transform the self-correction ability of large models from a heuristic process \cite{14} into a clear, trainable capability.
Our training approach is primarily reinforcement learning–based. During the reinforcement learning process, we specifically design a customized reward mechanism tailored for tool-calling scenarios, with a particular emphasis on multi-turn interactions. Concretely, the reward design encompasses multiple dimensions, including format reward, tool-name reward, parameter reward, and semantic reward of reflection, which together provide the model with multi-dimensional feedback and effectively guide its learning, and we further combine DAPO’s decoupled clipping range and dynamic sampling—expanding exploration while skipping near-zero-advantage rollouts—with GSPO’s sequence-level importance sampling and same-granularity clipping, which avoids token/sequence mismatch and stabilizes optimization. With this training methodology, our approach equips LLMs with genuine self-reflection and error-correction capabilities. On the BFCL v3 benchmark, our method yields significant improvements in LLM accuracy for multi-turn tool calling, thereby demonstrating its effectiveness in real-world applications.

We construct a Tool-Reflection-Bench based on the BUTTON dataset \cite{button} style. First, we collected tool-call failure cases from real-world scenarios and various benchmarks, analyzing and summarizing several common failure patterns. Next, We selected several existing tool-call datasets \cite{toolbench,toolace} and randomly combined them according to the call style of the BUTTON dataset and introduced these failure patterns into the data, disrupting the originally correct call processes to generate failure cases. Finally, we meticulously designed a reflection process to repair these failures, resulting in successful tool calls.
The training set includes the complete process described above to train LLMs to achieve true self-correction capabilities, while the test set only contains the first two steps, used to evaluate the self-correction abilities of the LLMs.
By constructing the Tool-Reflection-Bench in this manner, combined with our custom reward mechanism for tool calling, we have made breakthroughs in LLMs' self-correction abilities during training. Particularly in multi-turn tool-calling scenarios, we observed significant improvements in accuracy. Through the reasoning process from failure to correction, LLMs can more effectively identify and learn from potential mistakes, thus enhancing the model's stability and robustness in interactions. This makes the agent's behavior more robust and powerful.

In summary, our contributions are as follows:
\begin{itemize}
    \item We introduce an explicit, trainable reflection process that diagnoses the cause of a failed tool call using prior evidence and proposes a corrected, executable call. This transforms the "from failure to repair" process from a heuristic method into a learnable action strategy, enabling LLMs to genuinely possess self-reflection and error-correction capabilities, thereby enhancing the agent's multi-turn interactions with users.
    \item We design a more effective reward mechanism for tool call, tailored for RL training, using a GRPO-style objective function. This approach employs multi-dimensional rewards—format executability, tool name accuracy, parameter correctness, and semantic consistency—to mitigate sparse rewards and propagate signals across multi-turn trajectories.
    \item We propose Tool-Reflection-Bench, which collects failure patterns from real interaction scenarios and benchmark datasets, injects perturbations into correct calls, and attaches a reflection process to repair the calls. This allows for training LLMs in their Self-Correction ability in tool-calling scenarios.
    \item Our method significantly improves the accuracy of multi-turn tool calls and the ability to recover from tool call errors, while maintaining competitive single-turn tool call performance. We validate this by experiments on BFCL v3 \cite{bfcl} and Tool-Reflection-Bench.
\end{itemize}

\section{Method}
\subsection{Tool-Reflection-Bench}
The construction of Tool-Reflection-Bench consists of the following steps: perturbation-based disruptions, positive samples transformations, and the reflection repair process. The original positive samples are derived from BUTTON \cite{button} transformations and self-constructed based on few-shot prompts. The entire benchmark is divided into a training set and a test set, with approximately 5,000 samples in the training set, in addition to the reflection-augmented data constructed as described above, the training set also contains a very small portion of original data drawn from BUTTON \cite{button} and XLAM \cite{xlam}. And around 1,000 samples in the test set, the test set is exclusively composed of perturbation-derived items and does not include raw, unperturbed positives from BUTTON or XLAM.

\begin{figure*}[t]
  \centering
\includegraphics[width=0.999\textwidth] {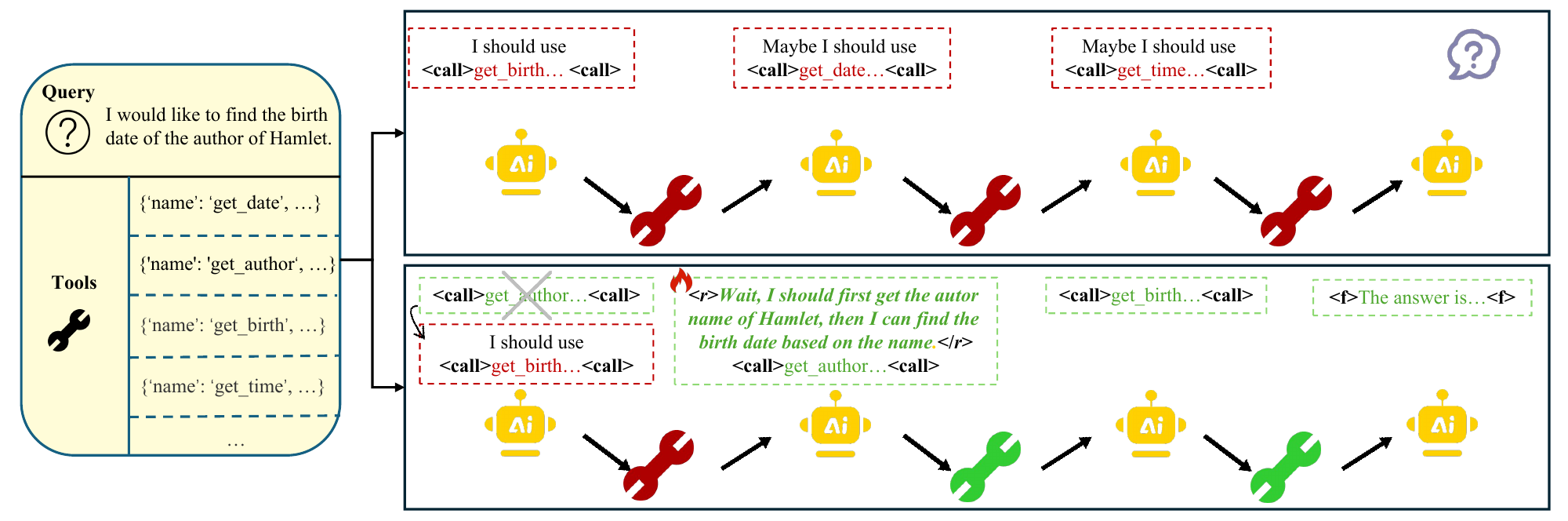}
  \caption{We illustrate the effectiveness of our method with an example. As shown in the figure, the left side presents the tool panel, while the upper-right part depicts industry-standard self-correction approaches, where models attempt to fix errors through heuristic trial-and-error reasoning or by relying on external feedback. In contrast, the lower-right part shows our approach: we introduce an explicit forced reflection process $<$r$>$, enabling the model to truly master the ability to repair errors based on its own failures.}
  \label{fig3}
\end{figure*}

\subsubsection{Perturbation-based Disruptions}
Let the initial correct message sequence be
\begin{equation}
\begin{split}
D^{+}
&= \Bigl(
m^{\mathrm{sys}}_{0},\, m^{\mathrm{usr}}_{1},\, m^{\mathrm{ast}}_{2},\, m^{\mathrm{tool}}_{3},\, \\
&\qquad m^{\mathrm{ast}}_{4},\, m^{\mathrm{tool}}_{5},\, \dots,\, m^{\mathrm{ast}}_{2k},\, m^{\mathrm{tool}}_{2k+1},\, \\
&\qquad \dots,\, m^{\mathrm{final}}_{n}
\Bigr).
\end{split}
\end{equation}
where $m^{\mathrm{sys}}_{0}$ is the system prompt, $m^{\mathrm{usr}}_{1}$ the user query, $m^{\mathrm{ast}}_{2i}$ the assistant's $i$-th tool call in structured form (e.g., \texttt{<call>[\{\dots\},\{\dots\},...]</call>}), $m^{\mathrm{tool}}_{2i+1}$ the tool return (JSON), and $m^{\mathrm{final}}_{n}$ the final answer.

We define a set of disruption operators
\begin{equation}
\mathcal{P}=\{P_{1},P_{2},P_{3},P_{4}\},
\end{equation}
each operating on an assistant call $m^{\mathrm{ast}}_{2k}$ and instantiating a common failure mode:
\begin{enumerate}
  \item $P_{1}$ \textbf{call-order swap}: replace the current tool call with the next-round tool call dialogue and force an error.
  \item $P_{2}$ \textbf{redundant call}: repeat the same tool at the step (unchanged/irrelevant arguments) and force an error.
  \item $P_{3}$ \textbf{missing call}: replace the intended tool by another tool and force an error.
  \item $P_{4}$ \textbf{argument error}: randomly corrupt the arguments of a call (missing/typed/alias/boundary) and force an error.
\end{enumerate}

These operators specify how a correct tool call can be broken.

\subsubsection{Positive Samples Transformations}
Given a clean trajectory $D^{+}$ and a chosen operator $P_{j}\in\mathcal{P}$ acting on step $2k$, we produce the negative (erroneous) context; no repair is performed in this step.
We construct the erroneous call
\begin{equation}
\tilde m^{\mathrm{ast}}_{2k}=\mathrm{ApplyPerturbation}\!\left(m^{\mathrm{ast}}_{2k},\,P_{j}\right),
\end{equation}
and simulate the tool's error feedback with a LLM $\mathcal{L}$:
\begin{equation}
\tilde m^{\mathrm{tool}}_{2k+1}= \mathcal{L}\!\left(\tilde m^{\mathrm{ast}}_{2k};\,\mathcal{L}\right).
\end{equation}
This yields the \textbf{negative} trajectory prefix
\begin{equation}
\begin{split}
D^{-}
&= \mathrm{Perturb}\!\left(D^{+},P_{j}\right) \\
&= \Bigl(
m^{\mathrm{sys}}_{0},\, m^{\mathrm{usr}}_{1},\, \ldots,\, \tilde m^{\mathrm{ast}}_{2k}, \\
&\qquad \tilde m^{\mathrm{tool}}_{2k+1}
\Bigr),
\end{split}
\end{equation}
which will later serve as evidence of failure. At this stage, the item consists only of the broken call and its error signal.

\subsubsection{Reflection Repair Process}
Given a clean trajectory $D^{+}$ and its perturbed prefix $D^{-}$, we present the LLM with a paired view of the step-$2k$ evidence:
\begin{equation}
\text{clean: }(m^{\mathrm{ast}}_{2k},\,m^{\mathrm{tool}}_{2k+1})
\quad\text{vs.}\quad
\text{broken: }(\tilde m^{\mathrm{ast}}_{2k},\,\tilde m^{\mathrm{tool}}_{2k+1}).
\end{equation}
The model outputs a response:
\begin{equation}
\langle\texttt{reflect}\rangle ref \langle/\texttt{reflect}\rangle ,
\end{equation}
where $ref$ briefly diagnoses the discrepancy, and $c$ proposes the fixed tool call. We then apply human supervision to obtain
$(ref^{\star},c^{\star})$, where $c^{\star}$ is set to the original correct call from the clean trajectory:
\begin{equation}
(ref,\,c)\;\xRightarrow[\text{human supervision}]{\text{post--editing}}\;(ref^{\star},\,c^{\star}).
\end{equation}
\begin{equation}
\mathcal{L}_{\Sigma}(c^{\star})=\texttt{Success Call}.
\end{equation}

\paragraph{Human supervision cost.}
Our supervision is performed at the \emph{trajectory} level. Specifically, we construct and retain approximately 5k multi-turn trajectories, and for each trajectory we only require post-editing the reflection text $ref$ (while $c^{\star}$ is directly copied from the clean call). Two annotators completed the post-editing process over 18 days, making the overall supervision cost controllable.

The finalized item is packaged as
\begin{equation}
x \;=\; \big(D^{-},\, ref^{\star},\, c^{\star},\, D^{+}_{>2k+1}\big),
\end{equation}
where $D^{+}_{>2k+1}$ is the untouched suffix of $D^{+}$ (including $m^{\mathrm{final}}_{n}$). We retain $x$ only if:
(i) tags/JSON are well-formed; (ii) $c^{\star}$ is executable; (iii) $ref^{\star}$ correctly cites the clean--broken contrast.

\subsection{Reward Design}
\label{sec:reward}

\paragraph{Preliminary.}
Given a model completion $C$ and a ground truth $G$, we decompose both into three (possibly empty) parts:
\begin{equation}
\begin{split}
C &\mapsto \bigl(c_{\text{ref}},\; C_{\text{calls}}=\{c_i\}_{i=1}^{m},\; c_{\text{final}}\bigr),\\
G &\mapsto \bigl(g_{\text{ref}},\; G_{\text{calls}}=\{g_j\}_{j=1}^{n},\; g_{\text{final}}\bigr).
\end{split}
\end{equation}

Here $c_{\text{ref}}$ is the diagnosis text wrapped in \texttt{<reflect></reflect>}, 
$C_{\text{calls}}$ is the \emph{multiset} of tool calls wrapped in one or more \texttt{<call></call>} blocks, 
and $c_{\text{final}}$ is the message wrapped in \texttt{<final></final>}. 
We treat $C_{\text{calls}}$ as a multiset because the same tool may be invoked multiple times within one completion, 
and we do not assume any canonical order when matching calls to the reference.
In our data format, each training example follows \emph{wrong call $\rightarrow$ reflection $\rightarrow$ corrected call}; 
the ground truth always contains a reflection and at least one valid tool call (i.e., $n\ge 1$), while the final message may be empty.
The ground truth can also be decomposed into the same three parts.

\begin{figure*}[htbp]
  \centering
  \begin{subfigure}{0.495\textwidth}
    \centering
    \includegraphics[width=\linewidth]{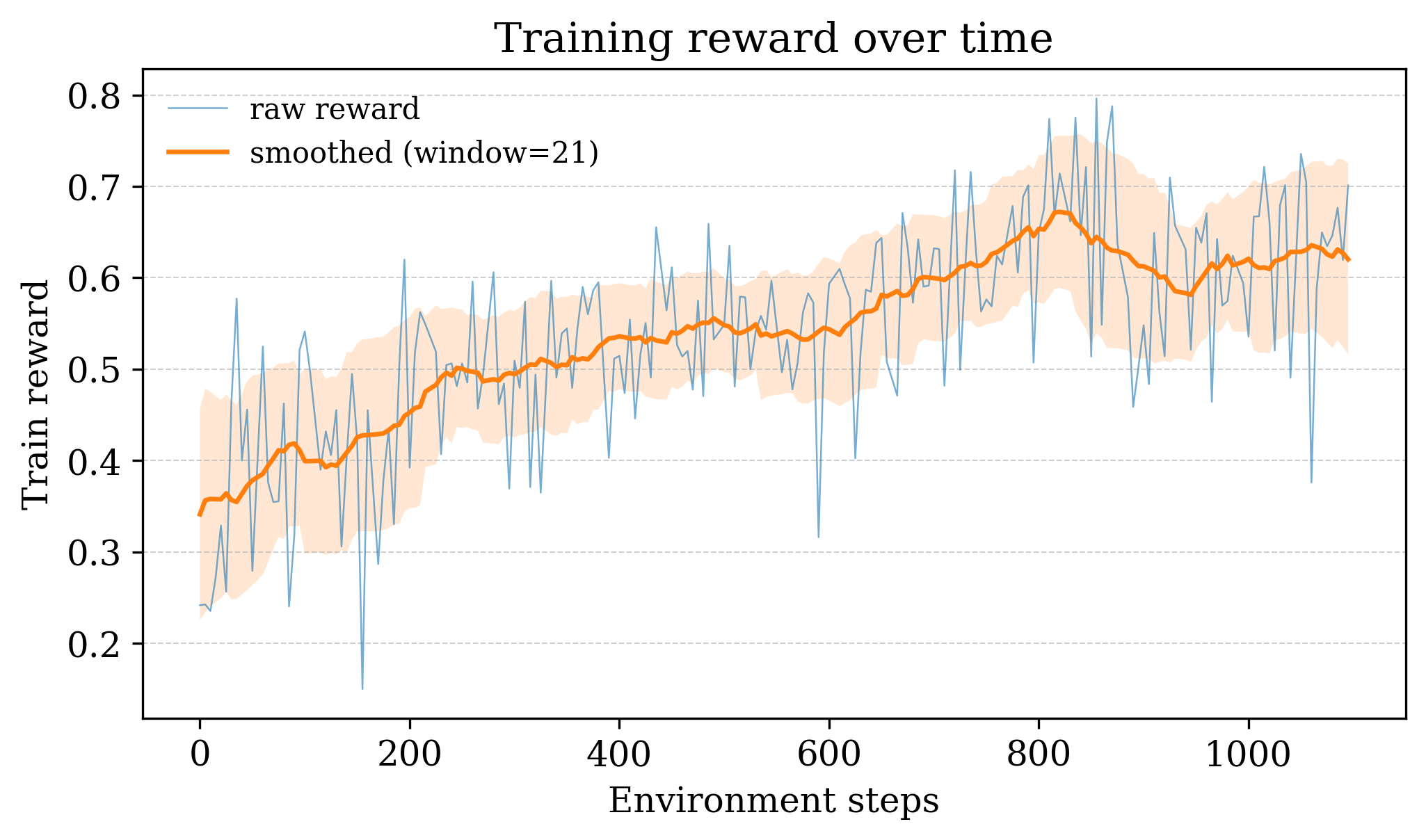}
    \caption{The reward curve of llama-3.1-8b-Instruct during RL training}\label{fig:bfcl}
  \end{subfigure}\hfill
  \begin{subfigure}{0.495\textwidth}
    \centering
    \includegraphics[width=\linewidth]{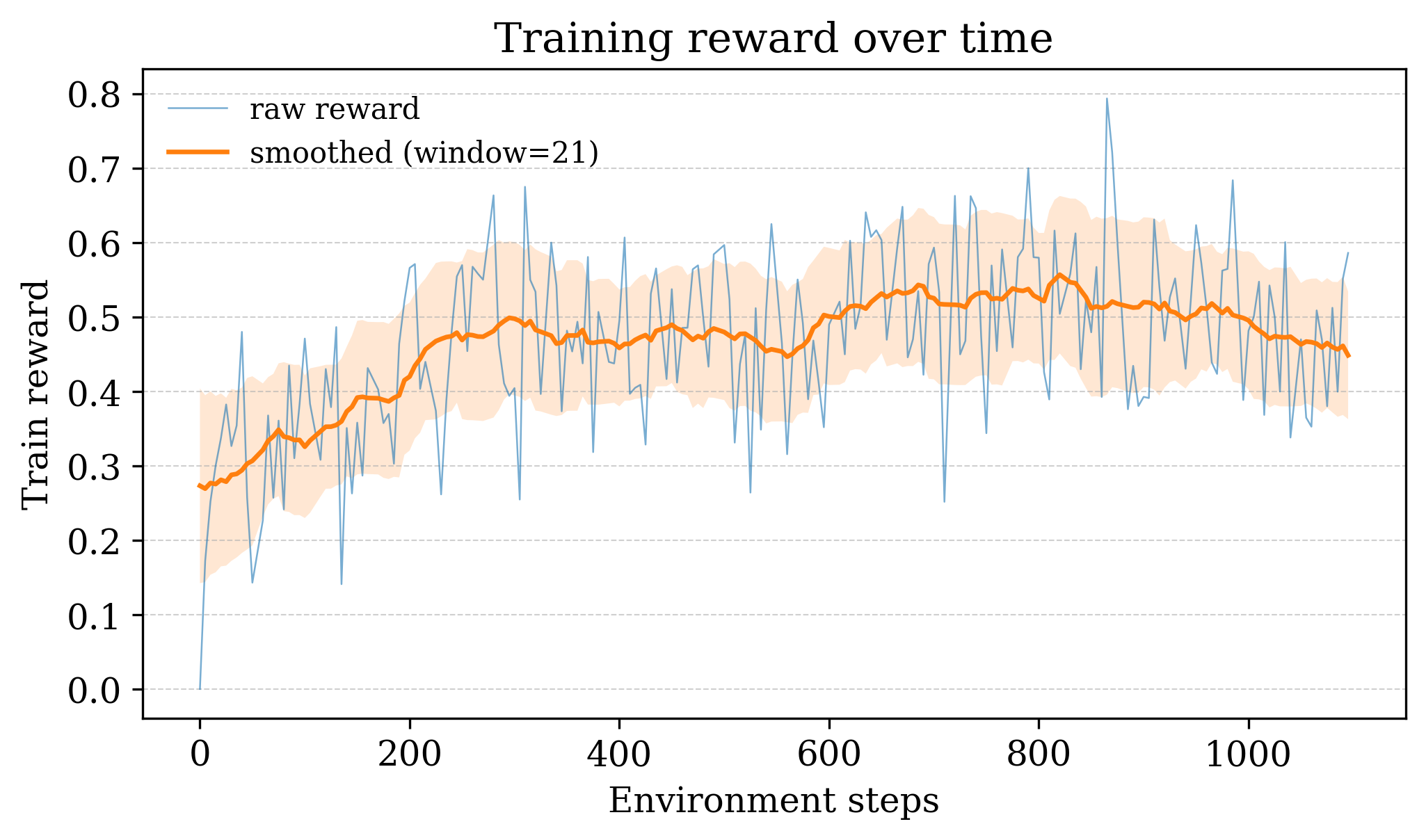}
    \caption{The reward curve of qwen2.5-7b-Instruct during RL training}\label{fig:trbench}
  \end{subfigure}
  \caption{The reward curves of llama-3.1-8B and Qwen2.5-7B during training, showing an overall upward trend.}
  \label{fig:two-rewards}
\end{figure*}

\paragraph{Component scores.}
We compute three component scores:
\begin{equation}
\begin{aligned}
s_{\text{ref}} \;&= \mathrm{Sim}(c_{\text{ref}}, g_{\text{ref}}), \\
s_{\text{call}} \;&= \mathbb{I}\!\left[\mathrm{EqualCalls}(C_{\text{calls}}, G_{\text{calls}})\right], \\
s_{\text{final}} \;&= \mathrm{Sim}(c_{\text{final}}, g_{\text{final}}),
\end{aligned}
\label{eq:comp-scores}
\end{equation}
where $\mathrm{Sim}\!\in[0,1]$ is a semantic similarity function, and
$\mathbb{I}[\cdot]$ is the indicator:
\begin{equation}
\mathbb{I}[P] \;=\;
\begin{cases}
1, & \text{if $P$ is true},\\
0, & \text{otherwise}.
\end{cases}
\end{equation}
We say $\mathrm{EqualCalls}(C_{\text{calls}}, G_{\text{calls}})$ holds iff the two multisets of calls can be put in a one-to-one correspondence such that for every matched pair the \textbf{tool name} is identical and the \textbf{argument} is identical.
We perform multiset matching: each produced call is matched to at most one reference call, and order is ignored.

\paragraph{Normalization with presence masks.}
Our goal is to keep the aggregated score in $[0,1]$ even when an instance specifies only a subset of targets
(e.g., only \texttt{<call>} without \texttt{<reflect>} or \texttt{<final>}). To this end we use normalization
to renormalize over the parts that actually appear in the ground truth, so the maximum remains $1$ regardless of how
many parts are present.

We define
\begin{equation}
\begin{aligned}
I_{\text{r}} \;&= \mathbb{I}\!\left[\,g_{\text{ref}}\neq\varnothing\,\right],\\
I_{\text{c}} \;&= \mathbb{I}\!\left[\,\lvert G_{\text{calls}}\rvert>0\,\right],\\
I_{\text{f}} \;&= \mathbb{I}\!\left[\,g_{\text{final}}\neq\varnothing\,\right].
\end{aligned}
\end{equation}
Let $(w_{\text{r}},w_{\text{c}},w_{\text{f}})\!\ge\!0$ be \emph{normalized} base weights (e.g., $w_{\text{r}}+w_{\text{c}}+w_{\text{f}}=1$).
We renormalize over the active parts via
\begin{equation}
W_{\text{act}} \;=\; w_{\text{r}} I_{\text{r}} + w_{\text{c}} I_{\text{c}} + w_{\text{f}} I_{\text{f}}.
\end{equation}
The aggregated structure/semantics score is then
\begin{equation}
S \;=\;\dfrac{w_{\text{r}} I_{\text{r}}\, s_{\text{ref}} \;+\; w_{\text{c}} I_{\text{c}}\, s_{\text{call}} \;+\; w_{\text{f}} I_{\text{f}}\, s_{\text{final}}}{W_{\text{act}}}.
\label{eq:S-agg}
\end{equation}
This normalization yields a \textbf{consistent} scoring standard across fully and partially supervised instances, avoiding artificial deflation of scores when some targets are absent.
 
\paragraph{Format/penalty factor.}
We designed structural penalties tailored for tool-call data formats. Specifically, $P_{miss}$ accounts for cases where the tool is not invoked at all, while $P_{extra}$ and $P_{count}$ penalize redundant calls and mismatches in the total number of calls, respectively. Let
\begin{equation}
n=|G_{\text{calls}}|,\qquad m=|C_{\text{calls}}|,
\end{equation}
Here $n$ and $m$ denote the number of tools invoked in the ground truth and completion calls.
Define the three components:
\begin{equation}
\begin{split}
P_{\text{miss}}
&= w_{\text{ref}}\;\mathbb{I}\!\left[\,g_{\text{ref}}\!\neq\!\varnothing \wedge c_{\text{ref}}\!=\!\varnothing\,\right] \\
&\quad +\; w_{\text{final}}\;\mathbb{I}\!\left[\,g_{\text{final}}\!\neq\!\varnothing \wedge c_{\text{final}}\!=\!\varnothing\,\right] \\
&\quad +\; w_{\text{calls}}\;\mathbb{I}\!\left[\,n>0 \wedge m=0\,\right].
\end{split}
\end{equation}

\begin{equation}
\begin{split}
P_{\text{extra}}
&= w_{\text{ref}}\;\mathbb{I}\!\left[\,c_{\text{ref}}\!\neq\!\varnothing \wedge g_{\text{ref}}\!=\!\varnothing\,\right] \\
&\quad +\; w_{\text{final}}\;\mathbb{I}\!\left[\,c_{\text{final}}\!\neq\!\varnothing \wedge g_{\text{final}}\!=\!\varnothing\,\right] \\
&\quad +\; w_{\text{calls}}\;\mathbb{I}\!\left[\,m>0 \wedge n=0\,\right].
\end{split}
\end{equation}

\begin{equation}
\begin{split}
P_{\text{count}}
&= w_{\text{calls}}\;\mathbb{I}\!\left[\,n>0 \wedge m>0 \wedge n\!\neq\!m\,\right] \\
&\quad \cdot \frac{\lvert n-m\rvert}{\max\{n,m\}}.
\end{split}
\end{equation}

\begin{table*}[!t]
\caption{Comparison across dimensions (Base, Miss\_Func, Miss\_Param, Long\_Context, Multi-turn Overall) on BFCL v3, all results in the table are reported as the average over three runs with different random seeds.}
\centering
\resizebox{0.95\textwidth}{!}{%
\begin{tabular}{l l c c c c c}
\toprule
\textbf{Models} & \textbf{Method} & \textbf{Base} & \textbf{Miss\_Func} & \textbf{Miss\_Param} & \textbf{Long\_Context} & \textbf{Multi-turn Overall} \\
\midrule
\multirow{2}{*}{Llama-3.1-8B-Instruct-FC}
  & Origin & 5.0 & 6.5 & 4.5 & 4.5 & 5.12 \\
  & \bb{Ours}   & \bb{9.5 ($\uparrow$95\%)} & \bb{7.0 ($\uparrow$8\%)} & \bb{5.0 ($\uparrow$11\%)} & \bb{7.0 ($\uparrow$56\%)} & \bb{7.12 ($\uparrow$39\%)} \\
\midrule
\multirow{2}{*}{Qwen2.5-7B-Instruct-FC}
  & Origin & 16.5 & 11.0 & 9.0  & 7.5 & 11.00 \\
  & \bb{Ours}   & \bb{22.0 ($\uparrow$33\%)} & \bb{13.0 ($\uparrow$18\%)} & \bb{13.5 ($\uparrow$50\%)} & \bb{11.0 ($\uparrow$47\%)} & \bb{14.88 ($\uparrow$35\%)} \\
\midrule
\multirow{2}{*}{Qwen3-4B-Instruct}
  & Origin & 18.0 & 19.0 & 13.5 & 14.5 & 16.25 \\
  & \bb{Ours}   & \bb{25.0 ($\uparrow$39\%)} & \bb{19.5 ($\uparrow$3\%)} & \bb{17.0 ($\uparrow$26\%)} & \bb{21.5 ($\uparrow$48\%)} & \bb{20.75 ($\uparrow$28\%)} \\
\bottomrule
\end{tabular}%
}
\label{tab:multi_turn_overall}
\end{table*}

Let $\mathrm{EqualCalls}$ be the schema-strict equality on bags of calls. And let $E$ = $\EC$
We use a reduction factor
\begin{equation}
r \;=\;
\begin{cases}
r_{\text{reduce}}, & \text{if } E,\\
1, & \text{else,}
\end{cases}\!,\quad
r_{\text{reduce}}\in(0,1].
\end{equation}
Aggregate the penalty as
\begin{equation}
P_{\text{total}} \;=\; P_{\text{miss}} \;+\; \beta_{\text{extra}}\,P_{\text{extra}} \;+\; \gamma_{\text{count}}\,P_{\text{count}},
\end{equation}
and define the instance-wise format factor. Let
$P_{\text{fmt}} := P_{\text{miss}} + P_{\text{extra}} + P_{\text{count}}$,
$[x]_{0}^{1} := \min\{1,\max\{0,x\}\}$, and
$F := \mathrm{FormatFactor}(C,G)$. Then
\begin{equation}
F =
\begin{cases}
1, & \text{if } P_{\text{fmt}}=0,\\
[1-\lambda_m P_{\text{total}} r]_{0}^{1}, & \text{otherwise.}
\end{cases}
\label{eq:format-final}
\end{equation}

Here $\beta_{\text{extra}},\gamma_{\text{count}},\lambda_m\ge 0$ control the strength of
the extra-part penalty, the count-mismatch penalty, and the overall scaling, respectively;
$(w_{\text{ref}},w_{\text{calls}},w_{\text{final}})\ge 0$ are part weights.

\paragraph{Core reward and backoff.}
The core reward is
\begin{equation}
R_{\text{core}} \;=\; S \cdot F.
\label{eq:Rcore}
\end{equation}
Early in training, $S$ contains a binary component ($s_{\text{call}}\!\in\!\{0,1\}$) and
$F$ applies hard penalties; small formatting or argument errors can drive $R_{\text{core}}$ close to zero.
This yields sparse or unstable gradients and large variance across samples. To stabilize learning and provide a
dense shaping signal when the exact-match objective is not yet achieved, we introduce a similarity backoff.
Let $[x]_{0}^{1}:=\max(0,\min(1,x))$ and
$S_{\text{b}}:=\mathrm{Sim}\!\big(\mathrm{concat}(C),\,\mathrm{concat}(G)\big)$:
\begin{equation}
R_{\text{total}}=
\begin{cases}
[\,R_{\text{core}}\,]_{0}^{1}, & R_{\text{core}} \ge \varepsilon,\\
[\,w_{\text{b}}\, S_{\text{b}}\,]_{0}^{1}, & \text{otherwise.}
\end{cases}
\label{eq:Rtotal}
\end{equation}
Here $w_{\text{b}}\in(0,1]$ and $\mathrm{concat}(\cdot)$ linearizes the messages.

\subsection{RL for Tool-Reflection-Bench}
\label{sec:rl}
We adopt a reinforcement-learning objective for tool calling that combines two complementary ideas:
\textbf{(i) DAPO-style decoupled clipping}~\cite{dapo}: we use a decoupled clipping range with different lower/upper bounds $(\varepsilon_{\text{low}},\varepsilon_{\text{high}})$ and a clip-higher policy (a looser upper bound when $r>1$ for positive advantages), and we skip uninformative prompt groups whose rollouts carry negligible learning signal;
\textbf{(ii) GSPO-style sequence-level importance sampling}~\cite{gspo}: we compute the importance ratio at the sequence level and apply clipping at the same granularity as the sequence-level reward, which avoids the mismatch between token-wise importance sampling and sequence-level rewards and stabilizes optimization.

\paragraph{Objective.}
Let $(q,a)$ denote the dialog context and the ground-truth targets, and let $\{o_i\}_{i=1}^{G}$ be $G$ candidates
sampled from the behavior policy $\pi_{\theta_{\text{old}}}(\cdot \mid q)$.
Each completion $o_i$ is scored by the reward in Sec.~\S\ref{sec:reward}, yielding $R_i\in[0,1]$.
We maximize a \emph{sequence-level, asymmetrically clipped} objective and minimize its negative as the loss:
\begin{equation}
\begin{aligned}
\mathcal{J}_{\text{RL}}(\theta)
&= \mathbb{E}\!\left[
\frac{1}{G}\sum_{i=1}^{G}
\min\!\big(r_i(\theta),\,\bar r_i(\theta)\big)\,\hat A_i
\right],\\
\bar r_i(\theta)
&:= \mathrm{clip}\!\big(r_i(\theta),\,1-\varepsilon_{\text{low}},\,1+\varepsilon_{\text{high}}\big).
\end{aligned}
\label{eq:rl-obj}
\end{equation}
Here the expectation is over $(q,a)\sim\mathcal{D}$ and $\{o_i\}\sim\pi_{\theta_{\text{old}}}(\cdot \mid q)$.
We define $\mathrm{clip}(x,a,b)=\min\{b,\max\{a,x\}\}$ and typically take
$\varepsilon_{\text{high}}>\varepsilon_{\text{low}}$ (``clip-higher'').

\paragraph{Prompt-group dynamic filtering.}
DAPO skips prompt groups whose candidates provide almost no learning signal (e.g., all-correct or all-wrong).
Concretely, let $\mu_R:=\mathrm{mean}(\{R_j\}_{j=1}^{G})$ and $\sigma_R:=\mathrm{std}(\{R_j\}_{j=1}^{G})$.
We define batch-normalized advantages and a group-level acceptance set:
\begin{equation}
\begin{aligned}
\hat A_i &= \frac{R_i-\mu_R}{\sigma_R},\\
\mathcal{S}(q,a) &= \Big\{\, i\in\{1,\ldots,G\} : |\hat A_i|>\tau_{\text{adv}} \,\Big\}.
\end{aligned}
\label{eq:adv}
\end{equation}
We further require sufficient reward dispersion within the group:
\begin{equation}
\begin{aligned}
\mathrm{Var}\!\big(\{R_i\}_{i=1}^{G}\big) &> \tau_{\text{var}},\\
0<|\mathcal{S}(q,a)| &< G.
\end{aligned}
\label{eq:dyn-filter}
\end{equation}
If \eqref{eq:dyn-filter} fails, we drop the zero-information rollouts and (optionally) draw up to $K$ additional
candidates from $\pi_{\theta_{\text{old}}}$, then re-apply the filter. Only indices in $\mathcal{S}(q,a)$ contribute
to the expectation in \eqref{eq:rl-obj}.

\paragraph{Sequence-level importance ratio.}
For a completion $o_i=(o_{i,1},\ldots,o_{i,|o_i|})$, we use the geometric-mean, length-normalized importance ratio:
\begin{equation}
r_i(\theta)
=
\left(
\prod_{t=1}^{|o_i|}
\frac{\pi_{\theta}\!\left(o_{i,t}\mid q,\,o_{i,<t}\right)}
     {\pi_{\theta_{\text{old}}}\!\left(o_{i,t}\mid q,\,o_{i,<t}\right)}
\right)^{\!1/|o_i|},
\label{eq:seq-ratio}
\end{equation}
and perform clipping at the same sequence granularity as the reward (see \eqref{eq:rl-obj}), thereby avoiding token/sequence granularity mismatch.

\section{Experiments}
\subsection{Experiment Settings}
In this part, we will detail the experimental setup, including datasets, hyperparameters, base models, and evaluation metrics.
\paragraph{Datasets.}
We conduct training on our self-constructed Tool-Reflection-Bench. After human supervision and post-editing, we retained approximately 5k samples in JSONL format to ensure compatibility with RL training under the Swift \cite{swift} framework.

\paragraph{Implementation Details.}
We train models for 1 epoch (a total of 1,000 steps) on 5,000 training samples, using the reward function defined in Sec.\ref{sec:reward}. For each training instance, 4 completions were sampled to form a group. The training parameters were set as follows: temperature = 0.85, repetition penalty = 1.1, epsilon = 0.2, epsilon-high = 0.28, with a dynamic sampling strategy adopted.

\paragraph{BFCL v3 Evaluation Metrics.}
We follow the official BFCL v3 multi-turn evaluation. Each subset contains 200 multi-turn conversations, and we report \textbf{conversation-level accuracy (pass@1)}: a conversation is correct only if it passes BFCL’s end-of-turn checks for all turns. BFCL executes the predicted tool calls and verifies correctness via both state-based (backend state) and response-based checks. We use the default termination rules (stop when no valid tool call is produced; force-terminate at 20 tool steps and mark incorrect). Multi-turn Overall is the unweighted average over the four subsets.

\paragraph{Tool-Reflection-Bench Evaluation Metrics.} To assess the model’s repair capability when tool calls fail, we used Tool-Reflection-Bench, with the evaluation metric being repair rate, Repair@n denotes that for the same data instance, if at least one out of n trials succeeds, the metric is recorded as 1; otherwise, it is 0.

\paragraph{Base Models.}
To verify the generalizability of Tool-Reflection-Bench and our training methodology, we conducted experiments using Llama3.1-8B \cite{llama3}, Qwen2.5-7B-Instruct \cite{qwen2_5}, and Qwen3-4B \cite{qwen3} as base models.

\begin{table}[!t]
\caption{Experimental Results of Open-Source and Closed-Source Models on the Tool-Reflection-Bench Test Set.}
\centering
\resizebox{\linewidth}{!}{%
\begin{tabular}{lccc}
\toprule
\textbf{Models} & \textbf{Repair@1 (\%)} & \textbf{Repair@3 (\%)} & \textbf{Repair@5 (\%)} \\
\midrule
\multicolumn{4}{c}{\textbf{Close-Sourced Models}}\\
\midrule
LongCat-Lite-8K-Chat & 2.3 & 3.4 & 4.9 \\
GPT-4o-mini          & 6.1  & 8.7  & 9.0  \\
GPT-4.1-mini         & 3.1  & 4.3  & 5.1  \\
\midrule
\multicolumn{4}{c}{\textbf{Open-Sourced Models}}\\
\midrule
Llama-3.1-8B-Instruct & 0.7 & 5.1 & 6.8 \\
Qwen2.5-7B-Instruct   & 2.4 & 6.1 & 8.0 \\
Qwen3-4B-Instruct     & 9.6 & 10.6 & 10.6 \\
\midrule
\multicolumn{4}{c}{\textbf{Open-Sourced Models Trained on Our Method}}\\
\midrule
\bb{Llama-3.1-8B-Instruct} & \bb{4.7}  & \bb{20.5} & \bb{26.4} \\
\bb{Qwen2.5-7B-Instruct}   & \bb{9.3}  & \bb{10.3} & \bb{11.4} \\
\bb{Qwen3-4B-Instruct}     & \bb{14.9} & \bb{18.5} & \bb{19.5} \\
\bottomrule
\end{tabular}%
}
\label{tab:repair_comparison}
\end{table}
\subsection{Experiment Results}

\begin{table*}[t]
\caption{Ablation study on BFCL v3 multi-turn (conversation-level Pass@1 accuracy, \%). We compare different RL variants on \textbf{Qwen2.5-7B-Instruct-FC}.}
\centering
\resizebox{0.92\textwidth}{!}{%
\begin{tabular}{l l c c c c c}
\toprule
\textbf{Base Model} & \textbf{RL Method} & \textbf{Base} & \textbf{Miss\_Func} & \textbf{Miss\_Param} & \textbf{Long\_Context} & \textbf{Overall} \\
\midrule
\multirow{4}{*}{Qwen2.5-7B-Instruct-FC}
& /    & 16.50 & 11.00 &  9.00 &  7.50 & 11.00 \\
& DAPO & 19.50 & 12.50 & 12.25 & 10.75 & 13.75 \\
& GSPO & 20.25 & 11.50 & 11.75 &  9.50 & 13.25 \\
& \bb{Ours} & \bb{22.00} & \bb{13.00} & \bb{13.50} & \bb{11.00} & \bb{14.88} \\
\bottomrule
\end{tabular}%
}
\label{tab:ablation_bfcl}
\end{table*}

\begin{table}[htbp]
\caption{Ablation study on Tool-Reflection-Bench (Repair@n, \%) for \textbf{Llama-3.1-8B-Instruct}. Prompt-only uses the inference prompt provided in the supplementary material (Appendix~\ref{sec:prompt_only}).}
\centering
\resizebox{\linewidth}{!}{
\begin{tabular}{l l c c c}
\toprule
\textbf{Base Model} & \textbf{Method} & \textbf{Repair@1} & \textbf{Repair@3} & \textbf{Repair@5} \\
\midrule
\multirow{4}{*}{Llama-3.1-8B-Instruct}
& /           & 0.70 &  5.10 &  6.80 \\
& Prompt Only & 1.50 &  9.30 & 12.60 \\
& SFT         & 3.20 & 12.40 & 16.90 \\
& \bb{RL}     & \bb{4.70} & \bb{20.50} & \bb{26.40} \\
\bottomrule
\end{tabular}}
\label{tab:ablation_trb}
\end{table}

\subsubsection{Result on BFCL v3}
\paragraph{Comparison with base models.}
Table~\ref{tab:multi_turn_overall} reports BFCL v3 multi-turn accuracy for the base models and our post-training.
Overall, our method consistently improves multi-turn tool-calling across all three backbones, with the largest gains on error patterns that require argument repair and retrieving missing information from long histories.
On \textbf{Llama-3.1-8B}, Base increases from 5.0 to 9.5 (+95\%) and Long\_Context from 4.5 to 7.0 (+56\%).
On \textbf{Qwen2.5-7B}, we observe the largest improvement on Miss\_Param (9.0 $\to$ 13.5, +50\%), indicating stronger parameter correction under tool feedback.
On \textbf{Qwen3-4B}, Multi-turn Overall increases from 16.25 to 20.75 (+28\%), accompanied by a substantial Long\_Context gain (+48\%).
In contrast, improvements on Miss\_Func are smaller (e.g., 19.0 $\to$ 19.5 on Qwen3-4B), suggesting that selecting the correct tool/function remains harder than repairing arguments given a plausible tool choice, which matches the focus of our reflection-driven repair training.

\subsubsection{Result on Tool-Reflection-Bench}
Tool-Reflection-Bench evaluates repair under tool-call failures; the test split consists solely of perturbation-derived failure cases (rather than clean trajectories), thus measuring recovery rather than memorization.
As shown in Table~\ref{tab:repair_comparison}, open-source baselines are weak at one try (Repair@1 $\leq$ 9.6\%) and improve only slightly with more attempts.
Our post-training consistently boosts repair across all backbones:
\textbf{Llama-3.1-8B-Instruct} improves from $0.7/5.1/6.8$ to \textbf{4.7/20.5/26.4},
\textbf{Qwen2.5-7B-Instruct} from $2.4/6.1/8.0$ to \textbf{9.3/10.3/11.4},
and \textbf{Qwen3-4B-Instruct} from $9.6/10.6/10.6$ to \textbf{14.9/18.5/19.5} (Repair@1/3/5).
The larger gains at Repair@3/5 suggest more robust reflection-to-repair behavior under repeated attempts.
Our fine-tuned models also outperform closed-source baselines (e.g., \textbf{LongCat-Lite-8K-Chat}~\cite{longcat}, \textbf{GPT-4o-mini}~\cite{4o,4os}, \textbf{GPT-4.1-mini}~\cite{41}) across $n\!\in\!\{1,3,5\}$.

\subsubsection{Ablation Studies}
\label{sec:ablation}

\paragraph{Analysis.}
Table~\ref{tab:ablation_bfcl} shows that both DAPO and GSPO improve over the base model on BFCL v3 multi-turn, while our full method achieves the best overall performance across all four subsets.
In particular, our gains are most pronounced on \textbf{Miss\_Param} and \textbf{Long\_Context}, consistent with our training signal that emphasizes reflection-driven argument repair and recovering missing information from long interaction histories.
Table~\ref{tab:ablation_trb} further validates the necessity of learning-based post-training for repair: prompt-only self-correction provides a limited improvement, SFT yields larger gains, and RL delivers the strongest repair capability, especially under multiple trials (Repair@3/5), indicating that the reward-driven optimization better aligns the model with robust correction behavior beyond imitation alone.

\section{Conclusion}
This paper proposes a structured reflection method for handling tool call failures, transforming the “from error to repair” process into an explicit, controllable, and trainable action. Our approach overcomes the limitations of previous heuristic, feedback-based self-correction methods in terms of controllability and stability.
We further construct Tool-Reflection-Bench for both training and evaluation, and design a task-specific reward function tailored to the tool-calling scenario. In the reinforcement learning stage, we combine the strengths of DAPO and GSPO to enhance training effectiveness.
Experimental results show that the proposed method significantly improves multi-turn tool call accuracy on BFCL v3 as well as error repair performance on Tool-Reflection-Bench. Overall, our method and dataset effectively enhance the reliability of tool interactions and offer a new perspective on enabling agents to acquire new capabilities by learning from failure.

\section{Limitations}
\label{sec:limitations}

\paragraph{Generalization beyond designed failures.}
Our training data is constructed via a set of perturbation-based disruptions that cover several common categories of tool-calling failures (e.g., missing/incorrect arguments and long-context dependencies). While our results on BFCL v3 multi-turn and Tool-Reflection-Bench demonstrate consistent improvements, these evaluations still represent a bounded set of failure modes. In real-world deployments, agents may encounter additional error types, such as ambiguous user intents, non-stationary tool APIs, partial tool outages, noisy tool outputs, or multi-tool plans with stronger temporal/causal dependencies. Extending perturbations and supervision to better approximate real failure distributions, as well as incorporating logs of naturally occurring failures, are promising directions.

\paragraph{Scaling to larger models.}
We validate our method on 4B--8B class models and observe substantial gains. However, the behavior of much larger models (e.g., 70B+) may differ: stronger base capabilities can reduce headroom, and the optimal balance between supervised signals and RL shaping may change. Evaluating and adapting the training recipe for larger backbones---including whether lighter post-training (e.g., SFT-only or fewer RL steps) suffices---remains future work.

\bibliography{custom}

@article{1,
  title={Function Calling in Large Language Models: Industrial Practices, Challenges, and Future Directions},
  author={WANG, MAOLIN and ZHANG, YINGYI and PENG, CUNYIN and CHEN, YICHENG and ZHOU, WEI and GU, JINJIE and ZHUANG, CHENYI and GUO, RUOCHENG and YU, BOWEN and WANG, WANYU and others},
  year={2025}
}

@article{2,
  title={Tool learning with large language models: A survey. CoRR abs/2405.17935(2024)},
  author={Qu, Changle and Dai, Sunhao and Wei, Xiaochi and Cai, Hengyi and Wang, Shuaiqiang and Yin, Dawei and Xu, Jun and Wen, J},
  journal={arXiv preprint arXiv:2405.17935},
  year={2024}
}

@article{3,
  title={Planning, creation, usage: Benchmarking llms for comprehensive tool utilization in real-world complex scenarios},
  author={Huang, Shijue and Zhong, Wanjun and Lu, Jianqiao and Zhu, Qi and Gao, Jiahui and Liu, Weiwen and Hou, Yutai and Zeng, Xingshan and Wang, Yasheng and Shang, Lifeng and others},
  journal={arXiv preprint arXiv:2401.17167},
  year={2024}
}

@article{toolbench,
  title={Toolllm: Facilitating large language models to master 16000+ real-world apis},
  author={Qin, Yujia and Liang, Shihao and Ye, Yining and Zhu, Kunlun and Yan, Lan and Lu, Yaxi and Lin, Yankai and Cong, Xin and Tang, Xiangru and Qian, Bill and others},
  journal={arXiv preprint arXiv:2307.16789},
  year={2023}
}

@article{5,
  title={Tool learning with large language models: A survey. CoRR abs/2405.17935(2024)},
  author={Qu, Changle and Dai, Sunhao and Wei, Xiaochi and Cai, Hengyi and Wang, Shuaiqiang and Yin, Dawei and Xu, Jun and Wen, J},
  journal={arXiv preprint arXiv:2405.17935},
  year={2024}
}

@article{6,
  title={Llm4eda: Emerging progress in large language models for electronic design automation},
  author={Zhong, Ruizhe and Du, Xingbo and Kai, Shixiong and Tang, Zhentao and Xu, Siyuan and Zhen, Hui-Ling and Hao, Jianye and Xu, Qiang and Yuan, Mingxuan and Yan, Junchi},
  journal={arXiv preprint arXiv:2401.12224},
  year={2023}
}

@article{7,
  title={Equipping language models with tool use capability for tabular data analysis in finance},
  author={Theuma, Adrian and Shareghi, Ehsan},
  journal={arXiv preprint arXiv:2401.15328},
  year={2024}
}

@article{8,
  title={Large language models can plan your travels rigorously with formal verification tools},
  author={Hao, Yilun and Chen, Yongchao and Zhang, Yang and Fan, Chuchu},
  journal={CoRR},
  year={2024}
}

@article{9,
  title={Sft or rl? an early investigation into training r1-like reasoning large vision-language models},
  author={Chen, Hardy and Tu, Haoqin and Wang, Fali and Liu, Hui and Tang, Xianfeng and Du, Xinya and Zhou, Yuyin and Xie, Cihang},
  journal={arXiv preprint arXiv:2504.11468},
  year={2025}
}

@article{10,
  title={Toolrl: Reward is all tool learning needs},
  author={Qian, Cheng and Acikgoz, Emre Can and He, Qi and Wang, Hongru and Chen, Xiusi and Hakkani-T{\"u}r, Dilek and Tur, Gokhan and Ji, Heng},
  journal={arXiv preprint arXiv:2504.13958},
  year={2025}
}

@article{11,
  title={Sparse rewards can self-train dialogue agents},
  author={Lattimer, Barrett Martin and Gangal, Varun and McDonald, Ryan and Yang, Yi},
  journal={arXiv preprint arXiv:2409.04617},
  year={2024}
}

@article{12,
  title={Torl: Scaling tool-integrated rl},
  author={Li, Xuefeng and Zou, Haoyang and Liu, Pengfei},
  journal={arXiv preprint arXiv:2503.23383},
  year={2025}
}

@article{13,
  title={Tl-training: A task-feature-based framework for training large language models in tool use},
  author={Ye, Junjie and Wu, Yilong and Li, Sixian and Yang, Yuming and Gui, Tao and Zhang, Qi and Huang, Xuanjing and Wang, Peng and Shi, Zhongchao and Fan, Jianping and others},
  journal={arXiv preprint arXiv:2412.15495},
  year={2024}
}

@article{14,
  title={Supercorrect: Supervising and correcting language models with error-driven insights},
  author={Yang, Ling and Yu, Zhaochen and Zhang, Tianjun and Xu, Minkai and Gonzalez, Joseph E and Cui, Bin and Yan, Shuicheng},
  journal={arXiv preprint arXiv:2410.09008},
  volume={9},
  year={2024}
}

@article{button,
  title={Facilitating multi-turn function calling for llms via compositional instruction tuning},
  author={Chen, Mingyang and Sun, Haoze and Li, Tianpeng and Yang, Fan and Liang, Hao and Lu, Keer and Cui, Bin and Zhang, Wentao and Zhou, Zenan and Chen, Weipeng},
  journal={arXiv preprint arXiv:2410.12952},
  year={2024}
}

@inproceedings{bfcl,
  title={The Berkeley Function Calling Leaderboard (BFCL): From Tool Use to Agentic Evaluation of Large Language Models},
  author={Patil, Shishir G and Mao, Huanzhi and Yan, Fanjia and Ji, Charlie Cheng-Jie and Suresh, Vishnu and Stoica, Ion and Gonzalez, Joseph E},
  booktitle={Forty-second International Conference on Machine Learning}
}

@article{toolace,
  title={Toolace: Winning the points of llm function calling},
  author={Liu, Weiwen and Huang, Xu and Zeng, Xingshan and Hao, Xinlong and Yu, Shuai and Li, Dexun and Wang, Shuai and Gan, Weinan and Liu, Zhengying and Yu, Yuanqing and others},
  journal={arXiv preprint arXiv:2409.00920},
  year={2024}
}

@article{xlam,
  title={xlam: A family of large action models to empower ai agent systems},
  author={Zhang, Jianguo and Lan, Tian and Zhu, Ming and Liu, Zuxin and Hoang, Thai and Kokane, Shirley and Yao, Weiran and Tan, Juntao and Prabhakar, Akshara and Chen, Haolin and others},
  journal={arXiv preprint arXiv:2409.03215},
  year={2024}
}

@article{19,
  title={FunReason: Enhancing Large Language Models' Function Calling via Self-Refinement Multiscale Loss and Automated Data Refinement},
  author={Hao, Bingguang and Wang, Maolin and Xu, Zengzhuang and Peng, Cunyin and Chen, Yicheng and Zhao, Xiangyu and Gu, Jinjie and Zhuang, Chenyi},
  journal={arXiv preprint arXiv:2505.20192},
  year={2025}
}

@article{apibank,
  title={Api-bank: A comprehensive benchmark for tool-augmented llms},
  author={Li, Minghao and Zhao, Yingxiu and Yu, Bowen and Song, Feifan and Li, Hangyu and Yu, Haiyang and Li, Zhoujun and Huang, Fei and Li, Yongbin},
  journal={arXiv preprint arXiv:2304.08244},
  year={2023}
}

@article{acebench,
  title={ACEBench: Who Wins the Match Point in Tool Learning?},
  author={Chen, Chen and Hao, Xinlong and Liu, Weiwen and Huang, Xu and Zeng, Xingshan and Yu, Shuai and Li, Dexun and Wang, Shuai and Gan, Weinan and Huang, Yuefeng and others},
  journal={arXiv e-prints},
  pages={arXiv--2501},
  year={2025}
}

@article{22,
  title={Self-reflection in llm agents: Effects on problem-solving performance},
  author={Renze, Matthew and Guven, Erhan},
  journal={arXiv preprint arXiv:2405.06682},
  year={2024}
}

@article{23,
  title={Large language models cannot self-correct reasoning yet},
  author={Huang, Jie and Chen, Xinyun and Mishra, Swaroop and Zheng, Huaixiu Steven and Yu, Adams Wei and Song, Xinying and Zhou, Denny},
  journal={arXiv preprint arXiv:2310.01798},
  year={2023}
}

@article{24,
  title={Self-Reflection Makes Large Language Models Safer, Less Biased, and Ideologically Neutral},
  author={Liu, Fengyuan and AlDahoul, Nouar and Eady, Gregory and Zaki, Yasir and Rahwan, Talal},
  journal={arXiv preprint arXiv:2406.10400},
  year={2024}
}

@article{25,
  title={Self-refine: Iterative refinement with self-feedback},
  author={Madaan, Aman and Tandon, Niket and Gupta, Prakhar and Hallinan, Skyler and Gao, Luyu and Wiegreffe, Sarah and Alon, Uri and Dziri, Nouha and Prabhumoye, Shrimai and Yang, Yiming and others},
  journal={Advances in Neural Information Processing Systems},
  volume={36},
  pages={46534--46594},
  year={2023}
}

@article{26,
  title={Large language models can self-correct with key condition verification},
  author={Wu, Zhenyu and Zeng, Qingkai and Zhang, Zhihan and Tan, Zhaoxuan and Shen, Chao and Jiang, Meng},
  journal={arXiv preprint arXiv:2405.14092},
  year={2024}
}

@article{27,
  title={PAG: Multi-Turn Reinforced LLM Self-Correction with Policy as Generative Verifier},
  author={Jiang, Yuhua and Xiong, Yuwen and Yuan, Yufeng and Xin, Chao and Xu, Wenyuan and Yue, Yu and Zhao, Qianchuan and Yan, Lin},
  journal={arXiv preprint arXiv:2506.10406},
  year={2025}
}

@article{28,
  title={Boosting LLM Reasoning via Spontaneous Self-Correction},
  author={Zhao, Xutong and Xu, Tengyu and Wang, Xuewei and Chen, Zhengxing and Jin, Di and Tan, Liang and Yu, Zishun and Zhao, Zhuokai and He, Yun and Wang, Sinong and others},
  journal={arXiv preprint arXiv:2506.06923},
  year={2025}
}

@article{29,
  title={Correcting Hallucinations in News Summaries: Exploration of Self-Correcting LLM Methods with External Knowledge},
  author={Vladika, Juraj and Soydemir, Ihsan and Matthes, Florian},
  journal={arXiv preprint arXiv:2506.19607},
  year={2025}
}

@inproceedings{30,
  title={Self-Reflective Retrieval-Augmented Generation (Self-RAG) in Analytical Systems},
  author={Saveliev, RI and Dendiuk, MV},
  booktitle={Forestry Education and Science: Current Challenges and Development Prospects. International Science-Practical Conference, October 23-25, 2024, Lviv, Ukraine},
  year={2024}
}

@article{31,
  title={Retool: Reinforcement learning for strategic tool use in llms},
  author={Feng, Jiazhan and Huang, Shijue and Qu, Xingwei and Zhang, Ge and Qin, Yujia and Zhong, Baoquan and Jiang, Chengquan and Chi, Jinxin and Zhong, Wanjun},
  journal={arXiv preprint arXiv:2504.11536},
  year={2025}
}

@article{dapo,
  title={Dapo: An open-source llm reinforcement learning system at scale},
  author={Yu, Qiying and Zhang, Zheng and Zhu, Ruofei and Yuan, Yufeng and Zuo, Xiaochen and Yue, Yu and Dai, Weinan and Fan, Tiantian and Liu, Gaohong and Liu, Lingjun and others},
  journal={arXiv preprint arXiv:2503.14476},
  year={2025}
}

@article{gspo,
  title={Group sequence policy optimization},
  author={Zheng, Chujie and Liu, Shixuan and Li, Mingze and Chen, Xiong-Hui and Yu, Bowen and Gao, Chang and Dang, Kai and Liu, Yuqiong and Men, Rui and Yang, An and others},
  journal={arXiv preprint arXiv:2507.18071},
  year={2025}
}

@inproceedings{swift,
  title={Swift: a scalable lightweight infrastructure for fine-tuning},
  author={Zhao, Yuze and Huang, Jintao and Hu, Jinghan and Wang, Xingjun and Mao, Yunlin and Zhang, Daoze and Jiang, Zeyinzi and Wu, Zhikai and Ai, Baole and Wang, Ang and others},
  booktitle={Proceedings of the AAAI Conference on Artificial Intelligence},
  volume={39},
  number={28},
  pages={29733--29735},
  year={2025}
}

@article{llama3,
  title={The llama 3 herd of models},
  author={Dubey, Abhimanyu and Jauhri, Abhinav and Pandey, Abhinav and Kadian, Abhishek and Al-Dahle, Ahmad and Letman, Aiesha and Mathur, Akhil and Schelten, Alan and Yang, Amy and Fan, Angela and others},
  journal={arXiv e-prints},
  pages={arXiv--2407},
  year={2024}
}

@article{qwen3,
  title={Qwen3 technical report},
  author={Yang, An and Li, Anfeng and Yang, Baosong and Zhang, Beichen and Hui, Binyuan and Zheng, Bo and Yu, Bowen and Gao, Chang and Huang, Chengen and Lv, Chenxu and others},
  journal={arXiv preprint arXiv:2505.09388},
  year={2025}
}

@article{qwen2_5,
  title={Qwen2. 5-coder technical report},
  author={Hui, Binyuan and Yang, Jian and Cui, Zeyu and Yang, Jiaxi and Liu, Dayiheng and Zhang, Lei and Liu, Tianyu and Zhang, Jiajun and Yu, Bowen and Lu, Keming and others},
  journal={arXiv preprint arXiv:2409.12186},
  year={2024}
}

@misc{4o,
  author       = {OpenAI},
  title        = {Hello GPT-4o},
  year         = {2024},
  month        = {May},
  howpublished = {\url{https://openai.com/index/hello-gpt-4o/}},
  note         = {Accessed: 2025-09-25}
}

@misc{4os,
  author        = {OpenAI},
  title         = {GPT-4o System Card},
  year          = {2024},
  eprint        = {2410.21276},
  archivePrefix = {arXiv},
  primaryClass  = {cs.CL},
  url           = {https://arxiv.org/abs/2410.21276},
  note          = {Accessed: 2025-09-25}
}

@misc{41,
  author       = {OpenAI},
  title        = {Introducing GPT-4.1 in the API},
  year         = {2025},
  month        = {April},
  howpublished = {\url{https://openai.com/index/gpt-4-1/}},
  note         = {Accessed: 2025-09-25}
}

@article{longcat,
  title={LongCat-Flash Technical Report},
  author={Team, Meituan LongCat and Li, Bei and Lei, Bingye and Wang, Bo and Rong, Bolin and Wang, Chao and Zhang, Chao and Gao, Chen and Zhang, Chen and Sun, Cheng and others},
  journal={arXiv preprint arXiv:2509.01322},
  year={2025}
}

@article{wang2024toolgen,
  title        = {ToolGen: Unified Tool Retrieval and Calling via Generation},
  author       = {Wang, Renxi and Han, Xudong and Ji, Lei and Wang, Shu and Baldwin, Timothy and Li, Haonan},
  journal      = {arXiv preprint arXiv:2410.03439},
  year         = {2024}
}

@inproceedings{xu2024enhancing,
  title        = {Enhancing Tool Retrieval with Iterative Feedback from Large Language Models},
  author       = {Xu, Qiancheng and Li, Yongqi and Xia, Heming and Li, Wenjie},
  booktitle    = {Findings of the Association for Computational Linguistics: EMNLP 2024},
  pages        = {9609--9619},
  year         = {2024}
}

@inproceedings{qu2025exploration,
  title        = {From Exploration to Mastery: Enabling LLMs to Master Tools via Self-Driven Interactions},
  author       = {Qu, Changle and Dai, Sunhao and Wei, Xiaochi and Cai, Hengyi and Wang, Shuaiqiang and Yin, Dawei and Xu, Jun and Wen, Ji-Rong},
  booktitle    = {International Conference on Learning Representations (ICLR)},
  year         = {2025}
}

\appendix

\section{Appendix}

\subsection{Use of LLMs}
This work leveraged LLMs to verify the mathematical soundness and symbolic accuracy of a few formulas in Sec.\ref{sec:theorm}.

\subsection{Related Works}
\subsubsection{Tool-augmented Large Language Models}
Integrating external tools into large language models has become a key approach to enhancing their functionality, surpassing the simple task of text generation. Traditional LLMs are limited by static knowledge, constrained to the data they were trained on. However, tool-augmented models extend the capabilities of LLMs by enabling them to interact with external resources \cite{xlam,19} (such as APIs \cite{apibank}, databases, and computational engines) through tool calls. This extension allows LLMs to access real-time data, perform external computations, and even interface with external hardware, making them more practical for solving complex real-world tasks that require dynamic information or specific external operations \cite{acebench}.
ToolBench \cite{toolbench} demonstrates the feasibility of integrating external tool calls into LLMs. Through such systems, LLMs can handle more specialized tasks. However, one major challenge of tool augmentation is how to effectively train LLMs to use these tools. Existing training methods, such as supervised fine-tuning and reinforcement learning, typically focus on optimizing single tool calls. This type of interaction often does not involve multi-turn tool calls or responses, which makes the limitations of current methods particularly apparent when errors occur during tool usage. In such cases, the model's ability to recover from errors becomes crucial.

\subsubsection{Self-Correction in LLMs}
Self-correction in large language models refers to the model's ability to diagnose its own errors and correct them based on previous actions \cite{23,24}. However, this area has not been fully explored. Existing self-correction techniques mostly rely on heuristic methods or unidirectional reasoning processes \cite{22}. 

Self-Refine framework \cite{25}, which involves having LLMs provide an initial response, followed by a reflection process where the model identifies flaws and makes improvements. Specifically, the same LLM acts as both the responder and the evaluator: the model first generates an initial response, then self-reflects and iteratively revises the output. This approach has been shown to enhance the performance of LLMs in certain domains. However, subsequent studies \cite{26,29} have found that relying solely on the model itself often fails to detect subtle errors. Some research \cite{27,28} has introduced auxiliary verifiers (such as additional models or mechanisms \cite{30,31}) to help check the correctness of the initial response. This external self-checking assistance avoids unnecessary repeated revisions, improving efficiency and enhancing the model's reasoning and verification capabilities. However, this approach remains highly sensitive to the specific phrasing of the prompts, with different prompt wordings leading to varying results \cite{24}.

Beyond prompt-only self-revision, recent work in tool-augmented agents improves reliability via interaction/execution feedback loops, e.g., unifying tool retrieval and calling via generation \cite{wang2024toolgen}, refining tool retrieval with iterative LLM feedback \cite{xu2024enhancing}, and trial-and-error frameworks that leverage tool interaction signals to iteratively improve tool understanding/documentation \cite{qu2025exploration}. 

However, even though these methods improve tool-use robustness, they primarily focus on leveraging online interaction signals (e.g., tool returns, success/failure) to drive behavior improvements. In contrast, our work targets self-correction itself as an explicit, trainable, and controllable capability: we formulate reflection-based error localization, diagnosis, and repair as a learning objective, and optimize it via supervised signals and reward shaping so that the model can reliably trigger and perform corrections during tool calling.

\subsection{Cost--Benefit Analysis on Tool-Reflection-Bench}
\label{sec:cost_benefit}

\begin{table*}[t]
\caption{\textbf{Cost--benefit comparison on Tool-Reflection-Bench} for \textbf{Llama-3.1-8B-Instruct}. We report Repair@\emph{n} (\%, higher is better) and training cost measured in wall-clock time on \textbf{4$\times$A100} GPUs. All RL variants are trained for \textbf{1000 steps} with identical data, sampling, and optimization settings; ``/'' denotes the untrained base model.}
\centering
\resizebox{\textwidth}{!}{%
\begin{tabular}{l l c c c c}
\toprule
\textbf{Base Model} & \textbf{RL Method} & \textbf{Repair@1} & \textbf{Repair@3} & \textbf{Repair@5} & \textbf{GPU Time (4$\times$A100)} \\
\midrule
\multirow{4}{*}{Llama-3.1-8B-Instruct}
& /    & 0.70 &  5.10 &  6.80 & / \\
& DAPO & 3.90 & 14.90 & 20.80 & 7h48m \\
& GSPO & 3.20 & 13.70 & 20.50 & 7h55m \\
& \bb{Ours} & \bb{4.70} & \bb{20.50} & \bb{26.40} & \bb{7h56m} \\
\bottomrule
\end{tabular}%
}
\label{tab:cost_benefit_trb}
\end{table*}

\paragraph{Analysis.}
Table~\ref{tab:cost_benefit_trb} quantifies the trade-off between RL complexity and performance under a fixed training budget.
Across DAPO, GSPO, and our method, the wall-clock cost is nearly identical (all $\approx$8 hours on 4$\times$A100 for 1000 steps), indicating that our pipeline does not introduce meaningful additional training overhead in this setting.
Despite comparable cost, our method achieves the best repair capability, improving Repair@5 from 20.8 (DAPO) / 20.5 (GSPO) to 26.4, and yielding a notably larger gain at higher $n$ (Repair@3/5), which suggests more reliable multi-try recovery when tool calls fail.

\subsection{Prompt for Prompt-only Self-correction}
\label{sec:prompt_only}

In this section, we provide a simplified prompt used for the \emph{Prompt Only} baseline in our ablation studies.
The prompt enforces a fixed output format (\texttt{<reflect>} + \texttt{<call>}) to encourage explicit diagnosis before proposing a corrected tool call.

\paragraph{How to run prompt-only self-correction}

\begin{promptbox}[System]
\textbf{Role.} You are a tool-calling assistant.

\textbf{Goal.} A previous tool call failed. Given the user goal, recent context, the failed tool call, and the tool error message,
produce (1) a concise reflection diagnosing the failure, and (2) exactly one corrected tool call that is executable under the provided tool schema.

\textbf{Inputs.} You will be given:
\begin{itemize}
  \item \texttt{[User Goal]}: the task the assistant is trying to accomplish.
  \item \texttt{[Tool API / Schema]}: function names and argument schemas.
  \item \texttt{[Context]}: recent messages and tool results.
  \item \texttt{[Failed Tool Call]}: the tool call that triggered the error.
  \item \texttt{[Tool Error / Feedback]}: the tool response/error message.
\end{itemize}

\textbf{Output format (strict).} Output \emph{must} follow this exact format with no extra text:
\begin{lstlisting}[style=promptlisting]
<reflect>
- Error type: (wrong tool / missing tool / wrong arguments / missing arguments / wrong values / redundant call / long-context mismatch).
- Evidence: cite the tool error and/or context that reveals the issue.
- Fix: state the minimal change needed to repair the call.
</reflect>
<call>{"name": "...", "arguments": {...}}</call>
\end{lstlisting}

\textbf{Constraints.}
\begin{enumerate}
  \item Produce exactly one \texttt{<reflect>} block and exactly one \texttt{<call>} block.
  \item The \texttt{<call>} must be valid JSON and executable under the provided schema.
  \item Do \emph{not} output \texttt{<final>...</final>} or any additional natural-language response.
  \item Prefer minimal edits: only change what is necessary to fix the error.
\end{enumerate}
\end{promptbox}

\begin{promptbox}[User]
\begin{lstlisting}[style=promptlisting]
[User Goal]
{USER_GOAL}

[Tool API / Schema]
{TOOL_SCHEMA_OR_DOCS}

[Context (recent messages and tool results)]
{CONTEXT}

[Failed Tool Call]
<call>{FAILED_CALL_JSON}</call>

[Tool Error / Feedback]
{ERROR_MESSAGE}
\end{lstlisting}
\end{promptbox}

\subsection{Prompt for Perturbation-based Disruptions}
In this section, we provide simplified prompts for generating the four types of tool call perturbations, enabling the community to reproduce our setting. The full prompts and implementation code will be released upon the paper’s acceptance.

\subsubsection{Prompt for Call-Order Swap}
\paragraph{How to construct an error tool call example}

\begin{promptbox}[System]
\textbf{Goal.} Prepend a controlled erroneous \texttt{<call>} and a consistent tool-error message
before the first assistant message, so the model must diagnose and repair.

\textbf{Procedure.}
\begin{enumerate}
  \item \emph{Extract calls:} Traverse messages and collect all assistant \texttt{<call>...</call>} blocks (regex).
  \item \emph{Choose function name:} Parse the \emph{last} call’s JSON to get \texttt{"name"}; fall back to a regex if needed.
  \item \emph{Synthesize wrong call (empty args):}
\begin{lstlisting}[style=promptlisting]
<call>[{"name":"<FUNC_FROM_LAST_CALL>","arguments":{}}]</call>
\end{lstlisting}
  \item \emph{Fabricate tool error (pretty JSON string):}
\begin{lstlisting}[style=promptlisting]
{"tool":"<FUNC_FROM_LAST_CALL>","status":"warning",
 "message":"The called function executed but returned partial/mismatched data because the arguments did not match the expected schema for this call.",
 "result": null}
\end{lstlisting}
  \item \emph{Insert pair:} Place the wrong assistant call and the tool error \emph{before} the original first assistant message.
  \item \emph{Elicit reflection:} Query the LLM with the System/User prompts above to obtain the reflection text,
  then prepend \texttt{<reflect>...</reflect>} to the original assistant message (the original correct call remains).
\end{enumerate}

\textbf{Notes.} Using the last call's function ensures schema plausibility; empty arguments induce a controlled failure; the synthetic tool message supplies concrete evidence for the subsequent reflection and repair.
\end{promptbox}

\paragraph{How to generate a reflection}

\begin{promptbox}[System]
You are an AI assistant that analyzes failed tool calls and provides reflective summaries.
Given an original tool call and a fabricated error response, generate a brief reflection
explaining why the call likely failed and how to correct it. Be concrete and concise.
\end{promptbox}

\begin{promptbox}[User]
Fill the placeholders \texttt{\{\{...\}\}} exactly.

\textbf{Original tool call:}
\begin{lstlisting}[style=promptlisting]
{{ORIGINAL_CALL}}
\end{lstlisting}

\textbf{Error response:}
\begin{lstlisting}[style=promptlisting]
{{FAKE_RESPONSE}}
\end{lstlisting}

Please provide a short reflection on the failure cause and the corrective action.
\end{promptbox}

\paragraph{An Example}
\begin{promptbox}[User]
\textbf{Original tool call:}
\begin{lstlisting}[style=promptlisting]
<call>[{"name":"searchArtistsByArtStyle","arguments":{}}]</call>
\end{lstlisting}

\textbf{Error response:}
\begin{lstlisting}[style=promptlisting]
{"tool":"searchArtistsByArtStyle","status":"warning",
 "message":"The called function executed but returned partial/mismatched data because the arguments did not match the expected schema for this call.",
 "result": null}
\end{lstlisting}

Please provide a brief reflection on why this tool call failed and what could be improved.
Keep it concise and helpful.
\end{promptbox}

\subsubsection{Prompt for Redundant Call}

\paragraph{How to construct a redundant tool call example}

\begin{promptbox}[System]
\textbf{Goal.} Inject a \emph{redundant} tool call inside an existing \texttt{<call>} list and a
matching redundant tool response, so the agent must identify and remove the duplication.

\textbf{Procedure.}
\begin{enumerate}
  \item \emph{Extract calls:} Traverse the dialogue and collect all assistant-side
        \texttt{<call>...\allowbreak</call>} blocks (regex).
  \item \emph{Pick a target (not the first):} Uniformly sample an assistant call position from
        $\{2,\ldots,|{\cal C}|\}$.
  \item \emph{Duplicate within the list:} Parse the target call's JSON.
        If it is a list, append a deep-copied first element; if it is a single dict, make a two-element list
        by duplicating it.
  \item \emph{Fabricate a redundant tool response:} Parse the following \texttt{tool} message.
        Duplicate its first item (or the dict itself) and mark it as redundant, e.g.
\begin{lstlisting}[style=promptlisting]
{"status":"redundant","message":"This item duplicates a previous result."}
\end{lstlisting}
  \item \emph{Keep the ground-truth call:} The \emph{correct} call is the original (non-duplicated)
        first element of the target call list.
  \item \emph{Place the repair evidence:} After the redundant \texttt{tool} message, insert an assistant
        message with \texttt{<reflect>} diagnosing the redundancy and a correct \texttt{<call>} (the
        non-duplicated one), followed by a \emph{clean} tool response (the original, without the redundant copy).
\end{enumerate}

\textbf{Notes.} This perturbation preserves schema but injects duplication at both call and response sides,
creating a realistic ``over-call'' pattern for reflection-and-repair.
\end{promptbox}

\paragraph{How to generate a reflection}

\begin{promptbox}[System]
You are an AI assistant that analyzes \emph{redundant} tool calls and provides reflective summaries.
Given a tool-call list and its redundant tool response, write a brief reflection that (i) identifies the
duplication, and (ii) states the correct next action (use only the necessary call with proper arguments).
Keep the reflection concise and actionable.
\end{promptbox}

\begin{promptbox}[User]
Fill the placeholders \texttt{\{\{...\}\}} exactly.

\textbf{Tool call list (after duplication):}
\begin{lstlisting}[style=promptlisting]
{{TOOL_CALL_LIST}}
\end{lstlisting}

\textbf{Redundant tool response:}
\begin{lstlisting}[style=promptlisting]
{{REDUNDANT_RESPONSE}}
\end{lstlisting}

Please provide a short reflection that points out the redundancy and explains how to proceed correctly.
\end{promptbox}

\paragraph{An Example}

\begin{promptbox}[User]
\textbf{Tool call list (after duplication):}
\begin{lstlisting}[style=promptlisting]
<call>[
  {"name":"searchArtistsByArtStyle","arguments":{"style":"impressionism"}},
  {"name":"searchArtistsByArtStyle","arguments":{"style":"impressionism"}}
]</call>
\end{lstlisting}

\textbf{Redundant tool response:}
\begin{lstlisting}[style=promptlisting]
[
  {"tool":"searchArtistsByArtStyle","status":"ok","items":[...]},
  {"tool":"searchArtistsByArtStyle","status":"redundant",
   "message":"This item duplicates a previous result.","items":[...]}
]
\end{lstlisting}

Please provide a brief reflection on why this redundant call occurred and how to proceed.
Keep it concise and helpful.
\end{promptbox}

\subsubsection{Prompt for Missing Call}

\paragraph{How to construct a missing-call perturbation example}

\begin{promptbox}[System]
\textbf{Goal.} Remove a necessary assistant \texttt{<call>} and make the subsequent call fail due to missing context, so the agent must \emph{recover the omitted call} and then proceed correctly.

\textbf{Procedure.}
\begin{enumerate}
  \item \emph{Extract calls:} Parse all assistant-side \texttt{<call>...</call>} blocks (regex).
  \item \emph{Select a removable call (not the last):} Uniformly sample an index $i \in \{1,\ldots,|{\cal C}|-1\}$.
  \item \emph{Find paired tool messages:} Locate the tool reply immediately after call $i$ (the one to remove), and the tool reply after call $i{+}1$ (the ``next'' call).
  \item \emph{Delete call $i$ and its tool reply.}
  \item \emph{Degrade the next call:} For the assistant \texttt{<call>} at (original) $i{+}1$, keep the function but set \texttt{"arguments":\{\}} (empty).
  \item \emph{Return an error for the next tool:} Replace that tool reply with an error JSON indicating ``missing required arguments''.
  \item \emph{Reflection and repair insertion:} After the error tool reply, insert:
  \begin{enumerate}
    \item an assistant message containing \texttt{<reflect>} that explains the omission and a \emph{reinstated} correct \texttt{<call>} (the removed call $i$);
    \item the original tool reply for the removed call $i$;
    \item the corrected next assistant call (its original, non-empty arguments);
    \item the corrected next tool reply (its original content).
  \end{enumerate}
\end{enumerate}

\textbf{Notes.} This perturbation creates a realistic ``missing prerequisite call'' failure: the subsequent step cannot execute without information from the omitted call. The reflection must (i) identify the omission and (ii) restore the correct call before proceeding.
\end{promptbox}

\paragraph{How to generate a reflection}

\begin{promptbox}[System]
You are an AI assistant that analyzes \emph{missing} tool calls and provides reflective summaries.
Given the omitted call (that should have been executed) and the resulting error response from the next step,
write a concise reflection that (i) identifies what was missing, and (ii) states how to proceed: first reinstate
the omitted call with correct arguments, then continue.
\end{promptbox}

\begin{promptbox}[User]
Fill the placeholders \texttt{\{\{...\}\}} exactly.

\textbf{Missing tool call (the one that should have been made):}
\begin{lstlisting}[style=promptlisting]
{{MISSING_CALL}}
\end{lstlisting}

\textbf{Error response (from the next step):}
\begin{lstlisting}[style=promptlisting]
{{ERROR_RESPONSE}}
\end{lstlisting}

Please provide a short reflection that explains the omission and the corrective sequence of actions.
\end{promptbox}

\paragraph{An Example}

\begin{promptbox}[User]
\textbf{Missing tool call:}
\begin{lstlisting}[style=promptlisting]
<call>[{"name":"fetchUserProfile","arguments":{"user_id":"u_1293"}}]</call>
\end{lstlisting}

\textbf{Error response (from the next step):}
\begin{lstlisting}[style=promptlisting]
[
  {"status":"error",
   "message":"Missing required arguments. The function call failed because necessary parameters were not provided.",
   "result": null}
]
\end{lstlisting}

Please provide a brief reflection on what was missing and how to proceed.
Keep it concise and helpful.
\end{promptbox}

\subsubsection{Prompt for Argument Error}

\paragraph{How to construct an argument–error perturbation example}

\begin{promptbox}[System]
\textbf{Goal.} Corrupt the arguments of an existing assistant \texttt{<call>} so that the paired tool reply returns a parameter–validation error, forcing the agent to \emph{diagnose mismatched/invalid arguments} and repair with the correct call.

\textbf{Procedure.}
\begin{enumerate}
  \item \emph{Extract calls:} Parse all assistant-side \texttt{<call>...</call>} blocks via regex.
  \item \emph{Select a call:} Uniformly sample one index \( i \in \{1,\ldots,|\mathcal{C}|\} \) and locate its immediate tool reply.
  \item \emph{Corrupt arguments:} Keep \texttt{"name"} unchanged; replace \texttt{"arguments"} with perturbed values (e.g., wrong types, out-of-range numbers, empty strings, unknown keys). The JSON stays well-formed:
\begin{lstlisting}[style=promptlisting]
<call>[{"name":"<FUNC_NAME>","arguments":{<WRONG_ARGS>}}]</call>
\end{lstlisting}
  \item \emph{Synthesize error reply:} Replace the paired tool message with a structured error indicating invalid parameters (e.g., \texttt{"error\_code":"INVALID\_PARAMETERS"} and an informative message).
  \item \emph{Reflection and repair insertion:} Immediately after the error, insert:
  \begin{enumerate}
    \item an assistant message with \texttt{<reflect>} that contrasts the wrong vs.\ correct arguments and states the fix;
    \item the \emph{original} (correct) call and its original (successful) tool reply.
  \end{enumerate}
\end{enumerate}

\textbf{Notes.} Do not alter the function name; only arguments are corrupted. Keep JSON/tags valid to isolate the failure mode to argument errors.
\end{promptbox}

\paragraph{How to generate a reflection}

\begin{promptbox}[System]
You are an AI assistant that analyzes incorrect tool-call \emph{parameters} and provides a reflective summary.
Given the correct call, the wrong call (with corrupted arguments), and the error response, write a brief reflection
that (i) pinpoints which arguments are incorrect and why, and (ii) states the corrected call.
Be concrete and concise.
\end{promptbox}

\begin{promptbox}[User]
Fill the placeholders \texttt{\{\{...\}\}} exactly.

\textbf{Correct tool call (ground truth):}
\begin{lstlisting}[style=promptlisting]
{{CORRECT_CALL}}
\end{lstlisting}

\textbf{Wrong tool call made (arguments corrupted):}
\begin{lstlisting}[style=promptlisting]
{{WRONG_CALL}}
\end{lstlisting}

\textbf{Error response:}
\begin{lstlisting}[style=promptlisting]
{{ERROR_RESPONSE}}
\end{lstlisting}

Please provide a short reflection that identifies the parameter issues and the corrective action.
\end{promptbox}

\paragraph{An Example}

\begin{promptbox}[User]
\textbf{Correct tool call:}
\begin{lstlisting}[style=promptlisting]
<call>[{"name":"bookFlight",
         "arguments":{"from":"SFO","to":"JFK","date":"2025-11-02","passengers":1}}]</call>
\end{lstlisting}

\textbf{Wrong tool call made:}
\begin{lstlisting}[style=promptlisting]
<call>[{"name":"bookFlight",
         "arguments":{"from":999999,"to":"","date":null,"passengers":"many"}}]</call>
\end{lstlisting}

\textbf{Error response:}
\begin{lstlisting}[style=promptlisting]
[{"status":"error",
  "message":"Parameter validation failed for bookFlight. One or more arguments are invalid.",
  "result": null,
  "error_code":"INVALID_PARAMETERS"}]
\end{lstlisting}

Please provide a brief reflection on which parameters are incorrect and how to fix them.
Keep it concise and helpful.
\end{promptbox}

\subsection{Training Data Case Study}
In this section, we extract one complete sample from each of the four perturbation modes for analysis. Due to space limitations, these four samples are provided in the supplementary material for reference, while here we only present a brief analysis of the data.

\subsubsection{Case Study of Call-Order Swap}

\paragraph{Setup.}
The user requests end–to–end logistics for a 10–person business trip (NYC\,$\rightarrow$\,MIA): search and \emph{book} round–trip flights, search and \emph{book} hotel rooms, and arrange airport--hotel ground transportation. The toolset exposes
\texttt{search\_flights}, \texttt{book\_flight}, \texttt{search\_hotels}, \texttt{book\_hotel}, and \texttt{arrange\_transportation}, where the last function \emph{depends} on confirmed itinerary details (arrival time and hotel).

\paragraph{Failure trace (order dependency violated).}
The agent correctly issues two discovery calls, then prematurely attempts to arrange transportation \emph{before} booking:
\begin{lstlisting}[style=promptlisting]
<tool_call>{"name":"search_flights", ...}</tool_call>
<tool_call>{"name":"search_hotels",  ...}</tool_call>

<tool_call>{"name":"arrange_transportation",
            "arguments":{"arrival_date":"2023-03-21",
                         "passenger_count":10,
                         "pickup_location":"MIA Airport",
                         "dropoff_location":"(to be decided)"}}</tool_call>
\end{lstlisting}
The tool returns a dependency error because \texttt{dropoff\_location} cannot be finalized prior to \texttt{book\_hotel} and pickup timing should be aligned to \texttt{book\_flight}:
\begin{lstlisting}[style=promptlisting]
Tool: {"status":"error",
       "message":"Transportation requires confirmed flight and hotel details.",
       "result": null}
\end{lstlisting}

\paragraph{Model reflection and repair.}
The agent emits a concise \texttt{<reflect>} that identifies the \emph{order–dependency} (transport must follow booking), then repairs by booking first and only then arranging transport:
\begin{lstlisting}[style=promptlisting]
<reflect>
Failure stems from an order dependency: transportation must follow confirmed
flight and hotel, as pickup/dropoff depend on the booked itinerary.
Correct plan: (1) book flight; (2) book hotel; (3) arrange transportation.
</reflect>

<tool_call>{"name":"book_flight", ...}</tool_call>
<tool_call>{"name":"book_hotel",  ...}</tool_call>
<tool_call>{"name":"arrange_transportation",
            "arguments":{"arrival_date":"2023-03-21",
                         "passenger_count":10,
                         "pickup_location":"MIA",
                         "dropoff_location":"Oceanview Hotel"}}</tool_call>
\end{lstlisting}

\paragraph{Why the reward favors the fix.}
Under our reward in Sec.~\S\ref{sec:reward}:
(i) the reflection receives a high semantic score $s_{\text{ref}}$ for explicitly diagnosing the \emph{order} constraint;
(ii) the repaired calls exactly match schema and parameters, yielding $s_{\text{call}}=1$ (order–\emph{invariant} matching across the multiset of calls), while the \emph{format/penalty factor} $F$ upweights the repaired trajectory because executability and dependency checks now pass;
(iii) the premature \texttt{arrange\_transportation} in the failed rollout triggers penalties in $F$ (dependency/executability), so $R_{\text{core}}$ is suppressed and the dynamic sampling filter (Sec.~\S\ref{sec:rl}) down–weights such uninformative rollouts.
Overall, the corrected plan yields a strictly larger $R_{\text{total}}$ than the erroneous one.

\paragraph{Summary.}
This example shows that explicit reflection converts a latent, long–range constraint (``book before transport'') into a short textual diagnosis that the policy can optimize against. Our order–invariant call equality avoids over–penalizing benign output permutations, while dependency–aware format shaping in $F$ enforces \emph{inter–call} preconditions. Empirically, such cases account for a substantial share of multi–turn gains, whereas single–turn metrics remain stable.

\subsubsection{Case Study of Redundant Call}

\paragraph{Setup and failure mode.}
The user requests the list of services for all NYC banks operating for $>10$ years. The agent first queries the bank roster with
\texttt{get\_operational\_banks}, then correctly queries services for \texttt{B001}, but erroneously \emph{repeats} the same service lookup for \texttt{B001} instead of proceeding to unseen banks (\texttt{B002--B004}). This is a canonical \emph{redundant call} error: wasted tool budget without adding new information.

\paragraph{Abbreviated trajectory.}
\begin{lstlisting}[style=promptlisting]
... 
<tool_call>
{"name":"get_operational_banks",
 "arguments":{"city":"New York City","min_years_operational":10}}
</tool_call>

User (tool): [{"name":"get_operational_banks","results":{"banks":[
  {"bank_id":"B001"}, {"bank_id":"B002"}, {"bank_id":"B003"}, {"bank_id":"B004"}]}}]

<tool_call>
{"name":"get_bank_services","arguments":{"bank_id":"B001"}}
</tool_call>

User (tool): [{"name":"get_bank_services","arguments":{"bank_id":"B001"},
 "results":{"services":[...]} }]

# Redundant (erroneous) repeat:
<tool_call>
{"name":"get_bank_services","arguments":{"bank_id":"B001"}}
</tool_call>

User (tool): [{"name":"get_bank_services","arguments":{"bank_id":"B001"},
 "results":{"services":[...], "note":"Duplicate request"}}]
...
\end{lstlisting}

\paragraph{Reflection and repair (ground truth).}
The reflection diagnoses duplication and prescribes de-duplication; the corrected plan proceeds with the remaining, unseen IDs:
\begin{lstlisting}[style=promptlisting]
<reflect>
The last call repeated B001, yielding redundant information. 
Avoid duplicates; continue with B002, B003, B004 exactly once each.
</reflect>

<tool_call>{"name":"get_bank_services","arguments":{"bank_id":"B002"}}</tool_call>
<tool_call>{"name":"get_bank_services","arguments":{"bank_id":"B003"}}</tool_call>
<tool_call>{"name":"get_bank_services","arguments":{"bank_id":"B004"}}</tool_call>
\end{lstlisting}

\paragraph{Why the model failed.}
The failure arises from (i) insufficient state tracking over the set of already-seen entities (here, bank IDs), and (ii) weak inductive bias against issuing calls whose \emph{marginal information gain} is near zero. In multi-turn settings, local myopic policies often re-issue the last successful pattern without cross-step deduplication.

\paragraph{How the reward steers recovery.}
Our scoring treats call sets as order-invariant but schema-strict; redundant calls trigger the count-mismatch component in the format factor $F$ (penalizing $|C_{\text{calls}}| \neq |G_{\text{calls}}|$) while \texttt{EqualCalls} fails due to multiset mismatch. The reflection text receives a positive semantic score if it explicitly identifies the duplication and prescribes the missing IDs, encouraging concise, actionable self-correction. Together, the structure score $S$ and format factor $F$ downweight redundant completions and upweight the repaired sequence.

\paragraph{Summary.}
This case shows that explicit reflection converts a silent efficiency bug into a supervised correction step: the agent (1) cites the duplicated identifier, (2) enumerates the remaining targets, and (3) completes them exactly once. Empirically, such reflection-shaped supervision reduces redundant tool usage and improves multi-turn success without harming single-turn accuracy.

\subsubsection{Case Study of Missing Call}

\paragraph{Setup.}
The user asks to register \emph{four} tax documents: (i) W-2 (ABC Corp), (ii) 1099-INT (First National Bank), (iii) property tax statement (county assessor), and (iv) Form~1098 (mortgage lender).
The tool schema exposes a single function \texttt{add\_tax\_documents({name, value, category, priority})} with \texttt{name,value} required.

\paragraph{Baseline failure (\textit{missing calls}).}
The baseline assistant emits only two \texttt{<tool\_call>}s (W-2, 1099-INT) and then stops, yielding a 50\% recall on required calls. Formally, let $G_{\text{calls}}$ contain the four intended calls and $C_{\text{calls}}$ the two produced calls. Then $|G_{\text{calls}}|=4$, $|C_{\text{calls}}|=2$, and the call–set equality test fails: $\mathrm{EqualCalls}(C_{\text{calls}},G_{\text{calls}})=0$. This is a typical \emph{missing-call} error in multi-item requests: the model recognizes the pattern “one item $\rightarrow$ one call” but truncates the sequence, leaving later items unprocessed.

\paragraph{Structured reflection (\textit{diagnosis}).}
Our method takes the partially executed trajectory as \emph{negative evidence} and the original request as \emph{positive intent} and generates an explicit reflection:
\begin{quote}\small
\texttt{<reflect>} “I missed 2 tool call(s). The user listed multiple items, and each item requires a separate call. I should enumerate all items and complete the remaining calls.” \texttt{</reflect>}
\end{quote}
The reflection correctly localizes the failure (under-counting of required calls), quantifies the deficit (missed$=2$), and states the repair rule (enumerate all items~$\Rightarrow$ one call per item).

\paragraph{Repairs (\textit{corrective calls}).}
Conditioned on the reflection, the agent appends the missing tool calls for the remaining items:
\begin{itemize}
\item \texttt{name:} Property tax statement; \texttt{value:} county assessor record; \texttt{category:} \texttt{personal};
\item \texttt{name:} Form 1098; \texttt{value:} mortgage interest statement; \texttt{category:} \texttt{personal}.
\end{itemize}
The assignments \texttt{work}$\rightarrow${W-2,1099-INT} and \texttt{personal}$\rightarrow${property tax, 1098} are semantically consistent: the former are employment/bank income records; the latter are household liabilities/taxes.\footnotesize\emph{(Any schema-compatible categorization would pass executability; ours also preserves natural semantics.)}\normalsize

\paragraph{Why this matters.}
This case highlights a frequent multi-turn brittleness: once the agent produces a plausible prefix of calls, it prematurely concludes and fails to cover all requested items. By making \textit{missingness} an explicit, trainable concept, structured reflection converts a sparse binary signal (success/failure) into actionable supervision:
\begin{enumerate}
\item \textbf{Detection:} Compare item cardinalities and arguments; compute $\mathbb{I}[|C_{\text{calls}}|<|G_{\text{calls}}|]$ and list uncovered entities.
\item \textbf{Diagnosis:} Attribute the error to \emph{enumeration/coverage} rather than formatting or parameters.
\item \textbf{Repair:} Synthesize the exact missing calls with schema-valid arguments; preserve already-correct calls.
\end{enumerate}

\paragraph{Summary.}
Empirically, such instances improve the model’s \emph{coverage discipline}: after training, we observe higher multi-item completion rates with negligible increase in redundant calls, indicating that the model learned “one-mentioned-item $\Rightarrow$ one-call” as a robust policy rather than overcalling.

\paragraph{Setup.}
The tool schema exposes multiple functions with \emph{schema--strict} parameters:
\begin{itemize}[leftmargin=1.2em,itemsep=1pt,topsep=2pt,parsep=0pt]
  \item \lstinline[style=promptlisting]$|check_plant_water_level(plant_location:string)|$
  \item \lstinline[style=promptlisting]$|start_watering(plant_location:string, duration:number)|$
  \item \lstinline[style=promptlisting]$|start_trimming(hedge_location:string)|$
  \item $\dots$
\end{itemize}
The user requests two primary actions in the backyard: (i) trim hedges and (ii) water all
potted plants for about 10 minutes; afterwards ensure plants have enough water and dispose clippings

\paragraph{Baseline failure (\textit{argument error}).}
The assistant issues
\begin{lstlisting}[style=promptlisting]
<call>[{"name":"check_plant_water_level","arguments":{}}]</call>
\end{lstlisting}
omitting the \emph{required} key \texttt{plant\_location}. The tool returns a schema warning that the
arguments ``did not match expected schema.'' Under our reward, the call-level indicator
$s_{\text{call}}$ is $0$ because the produced call fails schema equality (tool name matches, but the
argument map does not).

\paragraph{Structured reflection (\textit{diagnosis}).}
The reflection generated by our process states that the call ``failed because it did not include the required
arguments needed by the function's schema,'' and prescribes: ``ensure all necessary parameters are provided according
to the function's documentation.'' This localizes the error to \textbf{parameter mis-specification} (not tool
selection or ordering), and points to the concrete fix—satisfy the schema.

\paragraph{Repairs (\textit{efficient plan consistent with the request}).}
Given the user’s 10-minute target and the backyard scope, the corrected action set executes the two core operations
with schema-valid arguments:
\begin{itemize}[leftmargin=1.2em,itemsep=1pt,topsep=2pt,parsep=0pt]
  \item \lstinline[style=promptlisting]$|start_watering(plant_location="backyard", duration=10)|$
  \item \lstinline[style=promptlisting]$|start_trimming(hedge_location="backyard")|$
\end{itemize}
These can be dispatched in parallel (independent resources), achieving the requested time budget while ensuring
plants receive sufficient water and hedges are trimmed. This replaces the invalid pre-check with a direct,
time-bounded watering call that already satisfies the user’s constraint.

\paragraph{Why this matters.}
Argument errors are common in tool use and typically yield \emph{sparse} feedback (``schema mismatch''). By forcing the
model to (i) recognize the missing required field and (ii) restate the schema-conformant fix, the reflection step
converts a low-information error into actionable supervision. In our benchmark, such instances consistently improve:
\begin{enumerate}
  \item \textbf{Schema adherence:} higher exact-match rate on \texttt{name}/\texttt{arguments}.
  \item \textbf{Planning under constraints:} selection of parameterized calls (\texttt{duration=10}) aligned with user constraints instead of brittle pre-checks with empty arguments.
  \item \textbf{Stability:} fewer retries and warnings downstream because calls are executable on the first attempt.
\end{enumerate}

\paragraph{Summary.}
This case illustrates how reflection-guided repair turns a malformed \texttt{<call>} into a compact, correct, and time-efficient action plan.

\subsection{Test Data Case Study}
In this section, we present two representative test cases and their corresponding evaluation results as a case study, providing an intuitive demonstration of the effectiveness of our method and the model’s self-reflection capability for tool-call repair.
Since the original cases are relatively long, we include their full content in the supplementary material for reference and provide only the analysis here.

\subsubsection{Case I}

\paragraph{Setting.}
The tool set exposes three functions:
\texttt{getRecipes(max\_time, meal\_type)}, \texttt{getSmoothieIngredients(max\_time)}, and
\texttt{findComplementaryRecipes(recipes, ingredients)}.
The user asks for \emph{breakfast} recipes under 15 minutes and smoothie pairings under 5 minutes.

\paragraph{Failure mode (pre-training).}
The baseline model immediately issues
\begin{lstlisting}[style=promptlisting]
[{"name":"findComplementaryRecipes","parameters":{}}]
\end{lstlisting}
which violates the function schema (both \texttt{recipes} and \texttt{ingredients} are required).
The tool returns a schema-warning. Under our reward, this yields $s_{\text{call}}=0$ and triggers format penalties $F<1$
due to missing required parameters.

\paragraph{Reflection-driven repair (post-training).}
After RL on Tool-Reflection-Bench, the model first \emph{reflects} that the failure arises from absent inputs,
then correctly decomposes the task into \emph{produce inputs $\rightarrow$ compose}:
\begin{lstlisting}[style=promptlisting]
[{"name":"getRecipes","parameters":{"max_time":15,"meal_type":"breakfast"}}]
[{"name":"getSmoothieIngredients","parameters":{"max_time":5}}]
[{"name":"findComplementaryRecipes",
  "parameters":{"recipes": <from getRecipes>, "ingredients": <from getSmoothieIngredients>}}]
\end{lstlisting}
This satisfies the schema strictly (tool names and parameter maps match), making the call set correct and executable.

\paragraph{Why our method helps.}
(i) \textbf{Reward shaping:} The instance accrues a hard penalty when required fields are absent; after repair,
$s_{\text{call}}$ flips to $1$ and $F\!\to\!1$, raising $R_{\text{core}}=S\cdot F$ substantially. 
(ii) \textbf{Sequence-level RL:} The GSPO-style sequence-ratio with dual clipping aligns the optimization granularity with the sequence reward,
while DAPO-style dynamic filtering removes near-zero-advantage rollouts (all-wrong/all-correct), sharpening learning signals for this failure mode.

\paragraph{Takeaway.}
Compared to the baseline that \emph{jumps} to composition with empty inputs, the trained policy learns to (a) diagnose the schema error,
(b) \emph{stage} prerequisite calls to produce the missing inputs, and (c) complete the composition with a schema-valid call set.
This precisely matches our benchmark’s objective: enable robust, multi-turn tool use via reflection and repair.

\subsubsection{Case II}

\paragraph{Setting.}
Available tools include \texttt{get\_current\_season()}, \texttt{get\_seeds\_by\_season(season)}, 
\texttt{filter\_seeds\_by\_availability(seeds)}, \texttt{purchase\_seeds(seeds, quantity)}, and 
\texttt{calculate\_total\_cost(purchased\_seeds)}. 
The user asks to \emph{buy 10 packets of seasonal vegetable seeds} and \emph{report the total cost}.

\paragraph{Failure mode (pre-training).}
The baseline calls the aggregator first, with no inputs:
\begin{lstlisting}[style=promptlisting]
[{"name":"calculate_total_cost","parameters":{}}]
\end{lstlisting}
This violates the required schema (\texttt{purchased\_seeds} missing), producing a warning and yielding 
$s_{\text{call}}=0$ and a strong format penalty $F<1$ in our reward.

\paragraph{Reflection-driven repair (post-training).}
After RL on Tool-Reflection-Bench, the model first \emph{reflects} that costing requires purchased items, then
executes a staged pipeline to materialize prerequisites before aggregation:
\begin{lstlisting}[style=promptlisting]
[{"name":"get_current_season","parameters":{}}]
\end{lstlisting}
\begin{lstlisting}[style=promptlisting]
[{"name":"get_seeds_by_season","parameters":{"season":"<CUR_SEASON>"}}]
\end{lstlisting}
\begin{lstlisting}[style=promptlisting]
[{"name":"filter_seeds_by_availability","parameters":{"seeds":<SEASONAL_SEEDS>}}]
\end{lstlisting}
\begin{lstlisting}[style=promptlisting]
[{"name":"purchase_seeds","parameters":{"seeds":<AVAILABLE_SEEDS>,"quantity":10}}]
\end{lstlisting}
\begin{lstlisting}[style=promptlisting]
[{"name":"calculate_total_cost","parameters":{"purchased_seeds":<PURCHASED>}}]
\end{lstlisting}
Each call now matches tool name and parameter map exactly (schema-strict), so $s_{\text{call}}=1$ and $F\to 1$.

\paragraph{Why it works.}
\emph{Reward design} penalizes missing required fields and redundant structure, while granting full credit only when the 
\texttt{<call>} set exactly matches the ground truth (schema-strict, order-invariant). 
The \emph{sequence-level RL objective} (GSPO-style ratio, dual clipping) aligns optimization with sequence rewards, and 
\emph{DAPO-style dynamic filtering} removes near-zero-advantage groups, concentrating updates on informative failures.
Together these guide the policy to diagnose schema errors, stage prerequisite calls, and complete the costing correctly.

\paragraph{Takeaway.}
The trained policy no longer ``guesses'' totals from empty inputs. Instead, it \emph{plans $\rightarrow$ acquires data 
$\rightarrow$ purchases $\rightarrow$ aggregates}, a behavior precisely targeted by our reflection-and-repair rewards.

\subsection{Theoretical Analysis}
\label{sec:theorm}

We analyze the main design choices of our reward in Sec.~\S\ref{sec:reward} and the RL objective in Sec.~\S\ref{sec:rl}. Throughout, $\mathrm{Sim}\in[0,1]$, all
weights are nonnegative, presence masks are indicators, and $\mathrm{clip}(x,a,b)=\min\{b,\max\{a,x\}\}$. 
\emph{To avoid symbol overloading, we denote by $r_{\mathrm{fmt}}$ the format-penalty attenuation scalar used in Sec.~\S\ref{sec:reward} (called $r$ there), and by $r_{\mathrm{seq}}$ the sequence-level importance ratio in Sec.~\S\ref{sec:rl}.}

\subsubsection{Consistency of Presence-Mask Normalization}
Recall
\begin{equation}
\begin{aligned}
W_{\text{act}} &:= w_{\text{r}} I_{\text{r}} + w_{\text{c}} I_{\text{c}} + w_{\text{f}} I_{\text{f}},\\
S &:= \frac{w_{\text{r}} I_{\text{r}}\, s_{\text{ref}}
      + w_{\text{c}} I_{\text{c}}\, s_{\text{call}}
      + w_{\text{f}} I_{\text{f}}\, s_{\text{final}}}{W_{\text{act}}}.
\end{aligned}
\end{equation}

where $w_{\bullet}\!\ge 0$, $I_{\bullet}\!\in\{0,1\}$, at least one $I_{\bullet}=1$, $s_{\text{ref}},s_{\text{final}}\!\in[0,1]$, and $s_{\text{call}}\!\in\{0,1\}$.

\paragraph{Lemma 1 (Convex-combination form).}
Let $\mathcal{A}=\{k\in\{\text{r,c,f}\}: I_k=1\}$ and define
\begin{equation}
\alpha_k \;=\; \frac{w_k}{\sum_{j\in\mathcal{A}} w_j}
\quad\text{for }k\in\mathcal{A}.
\end{equation}
Then $\alpha_k\!\ge 0$, $\sum_{k\in\mathcal{A}}\alpha_k=1$, and
\begin{equation}
\begin{split}
S &= \sum_{k\in\mathcal{A}} \alpha_k\, s_k, \\
&\text{where } s_{\text{r}}=s_{\text{ref}},\;
s_{\text{c}}=s_{\text{call}},\;
s_{\text{f}}=s_{\text{final}}.
\end{split}
\end{equation}

\emph{Proof.}
Since $I_k=1$ iff $k\in\mathcal{A}$, the numerator equals $\sum_{k\in\mathcal{A}} w_k s_k$ and
$W_{\text{act}}=\sum_{k\in\mathcal{A}} w_k>0$. Divide both to obtain the stated form.

\paragraph{Proposition 1 (Boundedness, stability, and scale invariance).}
With $W_{\text{act}}>0$:
\begin{enumerate}
\item[\emph{(a)}] $S\in[0,1]$ and, more sharply, $S\in[\min_{k\in\mathcal{A}} s_k,\ \max_{k\in\mathcal{A}} s_k]$.
\item[\emph{(b)}] If one only toggles \emph{absent} parts (keeps $\mathcal{A}$ and $\{w_k\}_{k\in\mathcal{A}}$ unchanged), then $S$ is unchanged.
\item[\emph{(c)}] For any $\lambda>0$, replacing each active weight by $\lambda w_k$ leaves $S$ unchanged.
\end{enumerate}

\emph{Proof.}
(a) By Lemma~1, $S$ is a convex combination of $\{s_k\}_{k\in\mathcal{A}}$; the interval bound follows from $s_k\in[0,1]$.
(b) Absent-part toggles do not change $\mathcal{A}$ nor the active $w_k$. (c) Common scaling cancels in numerator/denominator.

\paragraph{Corollary 1 (Continuity and Lipschitzness).}
Fix $\mathcal{A}$ and $w_k$ for $k\in\mathcal{A}$. Then $S$ is an affine (hence continuous) map of $(s_k)_{k\in\mathcal{A}}$ with
\begin{equation}
|S-S'|\ \le\ \sum_{k\in\mathcal{A}} \alpha_k\,|s_k-s'_k|
\ \le\ \max_{k\in\mathcal{A}} |s_k-s'_k|,
\end{equation}
so $S$ is $1$-Lipschitz w.r.t.\ the $\ell_\infty$-norm on the active scores.

\paragraph{Remark.}
The definition via $\mathrm{clip}_{[0,1]}(\cdot)$ in \eqref{eq:Rtotal} is not needed for $S$ since the convex-combination form already implies $S\in[0,1]$.

\subsubsection{Format Factor: Boundedness, Monotonicity, and EqualCalls Attenuation}
Let
\begin{equation}
\begin{split}
P_{\text{total}}
&= P_{\text{miss}} + \beta_{\text{extra}}\,P_{\text{extra}} + \gamma_{\text{count}}\,P_{\text{count}},\\
&\text{where } \beta_{\text{extra}},\gamma_{\text{count}}\ge 0 \text{ and } P_{\bullet}\ge 0.
\end{split}
\end{equation}
and define the attenuation scalar
\begin{equation}
\begin{split}
\text{Let } E &:= \mathrm{EqualCalls}\!\big(C_{\text{calls}},\,G_{\text{calls}}\big).\\[-1pt]
r_{\mathrm{fmt}}
&=
\begin{cases}
r_{\text{reduce}}, & E,\\
1, & \text{otherwise,}
\end{cases}
\qquad r_{\text{reduce}}\in(0,1].
\end{split}
\label{eq:r-def}
\end{equation}
Consider
\begin{equation}
F \;=\; \mathrm{clip}_{[0,1]}\!\left( 1 - \lambda_m\, P_{\text{total}}\, r_{\mathrm{fmt}} \right),
\qquad \lambda_m \ge 0.
\label{eq:F-def}
\end{equation}
This is equivalent to the piecewise definition in \eqref{eq:format-final} since $P_{\text{miss}}{=}{P_{\text{extra}}}{=}{P_{\text{count}}}{=}0$ implies the inner value equals $1$.

\paragraph{Proposition 2 (Core properties of $F$).}
\begin{enumerate}
\item[\emph{(a)}] \emph{Boundedness and regularity.}
$F\in[0,1]$ for all inputs; $F$ is continuous, piecewise affine in $(P_{\text{miss}},P_{\text{extra}},P_{\text{count}})$ and $1$-Lipschitz w.r.t.\ its scalar argument before clipping.
\item[\emph{(b)}] \emph{Monotonicity.}
For fixed $(\lambda_m,r_{\mathrm{fmt}})$, $F$ is nonincreasing in $P_{\text{miss}},P_{\text{extra}},P_{\text{count}}$ and nonincreasing in $\lambda_m$ and in $r_{\mathrm{fmt}}$.
\item[\emph{(c)}] \emph{EqualCalls attenuation improves $F$.}
If $\mathrm{EqualCalls}$ holds so that $r_{\mathrm{fmt}}$ is replaced by $r_{\text{reduce}}\le 1$, then $F$ weakly increases.
\item[\emph{(d)}] \emph{Plateau characterization.}
$F=1$ iff $\lambda_m P_{\text{total}} r_{\mathrm{fmt}}=0$ (e.g., $P_{\text{total}}=0$ or $\lambda_m=0$). If $\lambda_m>0$ and $r_{\mathrm{fmt}}>0$, then $F=0$ iff $P_{\text{total}}\ge 1/(\lambda_m r_{\mathrm{fmt}})$.
\end{enumerate}

\paragraph{Corollary 2 (Sensitivity bound).}
Off the plateaus ($1-\lambda_m P_{\text{total}} r_{\mathrm{fmt}}\in(0,1)$),
\begin{equation}
\begin{split}
|\,\Delta F\,|
\ \le\ \lambda_m r_{\mathrm{fmt}}
\Big(
|\Delta P_{\text{miss}}|
+ \beta_{\text{extra}}|\Delta P_{\text{extra}}| \\
\qquad\qquad\qquad
+ \gamma_{\text{count}}|\Delta P_{\text{count}}|
\Big).
\end{split}
\end{equation}

\subsubsection{Core Reward with Similarity Backoff: Signal and Variance Control}
Let $R_{\text{core}}=S\cdot F$ as in \eqref{eq:Rcore}. To stabilize learning when $R_{\text{core}}$ is very small,
we introduce a similarity backoff. Define $[x]_{0}^{1}:=\min\{1,\max\{0,x\}\}$ and
$S_{\text{b}}:=\mathrm{Sim}\!\big(\mathrm{concat}(C),\,\mathrm{concat}(G)\big)$:
\begin{equation}
R_{\text{total}}=
\begin{cases}
[\,R_{\text{core}}\,]_{0}^{1}, & R_{\text{core}} \ge \varepsilon,\\
[\,w_{\text{b}}\,S_{\text{b}}\,]_{0}^{1}, & \text{otherwise.}
\end{cases}
\label{eq:Rtotal}
\end{equation}
Here $w_{\text{b}}\in(0,1]$ and $\varepsilon>0$. Note that $R_{\text{core}}\in[0,1]$ already, so clipping is redundant
but harmless, and it keeps the two branches notationally symmetric.

We analyze its effect under a standard policy-gradient estimator
$\nabla_\theta \mathbb{E}[R_{\text{total}}]
=\mathbb{E}\!\left[R_{\text{total}}\;\nabla_\theta \log \pi_\theta(\cdot)\right]$.

\paragraph{Lemma 3 (Uniform bounded variance of the reward).}
Since $R_{\text{total}}\in[0,1]$, we have $\mathrm{Var}(R_{\text{total}})\le \tfrac{1}{4}$ for any data distribution.

\paragraph{Lemma 4 (Non-degenerate gradient second moment on the backoff branch).}
Let $\mathcal{B}:=\{R_{\text{core}}<\varepsilon\}$ with $\mathbb{P}(\mathcal{B})=p>0$.
Let $g_\theta:=\nabla_\theta \log \pi_\theta(\cdot)$.
Assume $S_{\text{b}}\ge \sigma$ a.s.\ on $\mathcal{B}$ for some $\sigma>0$ and
$\mathbb{E}\!\left[\|g_\theta\|^2\,\mathbf{1}_{\mathcal{B}}\right]>0$.
Then
\begin{equation}
\begin{aligned}
\mathbb{E}\!\left[\ \big\|R_{\text{total}}\,g_\theta\big\|^2\ \right]
&\ge (w_{\text{b}}\sigma)^2\ \mathbb{E}\!\left[\ \|g_\theta\|^2\,\mathbf{1}_{\mathcal{B}}\ \right] \\
&> 0.
\end{aligned}
\end{equation}
\emph{Implication.} When $R_{\text{core}}$ frequently approaches $0$ (early in training),
the backoff branch prevents the gradient second moment from degenerating; together with the variance upper bound
in Lemma~3, this stabilizes the optimization updates.

\subsubsection{Sequence-Level Importance Sampling and Clipping}
Let the sampled completion be $o=(o_{1},\dots,o_{T})$, and define the sequence-level (geometric-mean, length-normalized) ratio:
\begin{equation}
\begin{split}
r_{\mathrm{seq}}(\theta)
&=
\left(
\prod_{t=1}^{T}
\frac{\pi_{\theta}\!\left(o_{t}\mid q,\,o_{<t}\right)}
     {\pi_{\theta_{\text{old}}}\!\left(o_{t}\mid q,\,o_{<t}\right)}
\right)^{\!1/T} \\
&=
\exp\!\left(\frac{1}{T}\sum_{t=1}^{T}\log \rho_t\right), \\
\rho_t
&:= \frac{\pi_{\theta}\!\left(o_t\mid q,\,o_{<t}\right)}
          {\pi_{\theta_{\text{old}}}\!\left(o_t\mid q,\,o_{<t}\right)}.
\end{split}
\label{eq:seq-ratio-fixed}
\end{equation}

\paragraph{Proposition 3 (Length-independent ratio range under bounded log-ratios).}
If $\log \rho_t\in[-L,L]$ a.s.\ for some $L>0$, then
\begin{equation}
e^{-L} \;\le\; r_{\mathrm{seq}}(\theta) \;\le\; e^{L} \qquad \text{for all } T\ge 1,
\end{equation}
whereas the unnormalized product ratio ranges in $[e^{-LT},e^{LT}]$.

\emph{Implication.}
The geometric mean aligns the \emph{ratio granularity} with the \emph{sequence-level reward} in \eqref{eq:rl-obj}, prevents exponential blow-up with $T$, and—together with dual clipping—reduces variance at the sequence level.

\subsubsection{Dynamic Filtering of Prompt Groups (DAPO-style)}
Let a prompt group produce $G$ rollouts $\{o_i\}_{i=1}^{G}$ with rewards $R_i\!\in[0,1]$ and batch $z$-scored
advantages
\begin{equation}
\begin{aligned}
\hat A_i &:= \frac{R_i-\bar R}{s_R},\\
\bar R &:= \frac{1}{G}\sum_{j=1}^{G} R_j,\\
s_R &:= \sqrt{\frac{1}{G}\sum_{j=1}^{G}(R_j-\bar R)^2}\;>\;0.
\end{aligned}
\end{equation}
Define the \emph{accepted} set
\begin{equation}
\begin{aligned}
\mathcal{S} &:= \big\{\, i : |\hat A_i|>\tau_{\text{adv}} \,\big\},\\
0<|\mathcal{S}| &< G,\\
\mathrm{Var}\!\big(\{R_i\}_{i=1}^{G}\big) &> \tau_{\text{var}} \;>\; 0.
\end{aligned}
\end{equation}

Write the per-sample (sequence-level, dual-clipped) PPO-like term as
\begin{equation}
\begin{aligned}
\ell_i(\theta)
&:= \min\!\Big(r_{\mathrm{seq},i}(\theta),\,\bar r_{\mathrm{seq},i}(\theta)\Big)\,\hat A_i,\\
\bar r_{\mathrm{seq},i}(\theta)
&:= \mathrm{clip}\!\Big(r_{\mathrm{seq},i}(\theta),\,1-\varepsilon_{\text{low}},\,1+\varepsilon_{\text{high}}\Big).
\end{aligned}
\end{equation}
and denote its gradient by $g_i(\theta)=\nabla_\theta \ell_i(\theta)$.
Assume the usual score-function bound and clipped ratio range:
\begin{equation}
\begin{aligned}
\big\|\nabla_\theta \log \pi_\theta(o_{i,t}\mid q,o_{i,<t})\big\|
&\le B_\pi,\\
r_{\mathrm{seq},i}(\theta)
&\in\big[\,1-\varepsilon_{\text{low}},\,1+\varepsilon_{\text{high}}\,\big].
\end{aligned}
\tag{$\star$}
\end{equation}

\paragraph{A uniform bound on per-rollout gradients.}
Since $r_{\mathrm{seq},i}(\theta)$ is the geometric mean of token ratios, let $T_i:=|o_i|$ and $h_{i,t}:=(q,o_{i,<t})$.
Then
\begin{equation}
\begin{aligned}
\nabla_\theta r_{\mathrm{seq},i}(\theta)
&= r_{\mathrm{seq},i}(\theta)\,\frac{1}{T_i}
\sum_{t=1}^{T_i} \nabla_\theta \log \pi_\theta(o_{i,t}\mid h_{i,t}) \\
&= \frac{r_{\mathrm{seq},i}(\theta)}{T_i}
\sum_{t=1}^{T_i} \nabla_\theta \log \pi_\theta(o_{i,t}\mid h_{i,t}).
\end{aligned}
\end{equation}
Using $(\star)$ and that the clipped branch is constant on plateaus, there exists a finite
$C_\psi= (1+\varepsilon_{\text{high}})\,B_\pi$ such that
\begin{equation}
\|g_i(\theta)\| \;\le\; C_\psi\,|\hat A_i|
\qquad \text{for all } i,\ \theta.
\label{eq:gibound}
\end{equation}

\paragraph{Lemma 5 (Zero or near-zero advantages).}
\begin{enumerate}[label=(\alph*),leftmargin=1.5em,itemsep=0pt,topsep=2pt]
\item If $\hat A_i=0$, removing $o_i$ leaves the group-wise expected gradient unchanged.
\item If $|\hat A_i|\le \tau_{\text{adv}}$, then for any $\theta$,
\begin{equation}
\begin{aligned}
\big\|\,\mathbb{E}[\,g_i(\theta)\,]\,\big\| &\le C_\psi\,\tau_{\text{adv}},\\
\mathbb{E}\!\left[\ \|g_i(\theta)\|^2\ \right] &\le C_\psi^2\,\tau_{\text{adv}}^{\,2}.
\end{aligned}
\end{equation}
\end{enumerate}

\emph{Proof.}
(a) The contribution is proportional to $\hat A_i$. (b) Apply \eqref{eq:gibound} and take expectations.

\paragraph{Bias and variance effects with $\frac{1}{G}$ normalization.}
Let the \emph{filtered} group gradient be
\begin{equation}
\begin{aligned}
\tilde g(\theta) &:= \frac{1}{G}\sum_{i\in\mathcal{S}} g_i(\theta),\\
g(\theta) &:= \frac{1}{G}\sum_{i=1}^{G} g_i(\theta)\qquad\text{(unfiltered)}.
\end{aligned}
\end{equation}

Define the discarded set $\mathcal{S}^c=\{1,\dots,G\}\setminus\mathcal{S}$.
Then
\begin{align}
\mathbb{E}[\tilde g(\theta)]-\mathbb{E}[g(\theta)]
&=\; -\,\frac{1}{G}\sum_{i\in\mathcal{S}^c}\mathbb{E}[\,g_i(\theta)\,],\\
\Big\|\ \mathbb{E}[\tilde g(\theta)]-\mathbb{E}[g(\theta)]\ \Big\|
&\le\; \frac{|\mathcal{S}^c|}{G}\,C_\psi\,\tau_{\text{adv}}
\;\le\; C_\psi\,\tau_{\text{adv}}, \nonumber
\end{align}
using Lemma~5(b). Moreover,
\begin{equation}
\mathbb{E}\!\left[\ \Big\|\frac{1}{G}\sum_{i\in\mathcal{S}^c} g_i(\theta)\Big\|^2\ \right]
\;\le\; \frac{|\mathcal{S}^c|}{G^2}\,C_\psi^{\,2}\,\tau_{\text{adv}}^{\,2},
\end{equation}
thus, discarding near-zero advantageous terms induces at most an $O(\tau_{\text{adv}}^{\,2})$-level change in the second moment; with respect to the $\frac{1}{G}$ normalization, it does not introduce any additional scaling bias.

\paragraph{Acceptance constraints avoid degeneracy.}
The constraints $0<|\mathcal{S}|<G$ and $\mathrm{Var}(\{R_i\})>\tau_{\text{var}}$ ensure: (i) the batch standardization
$s_R$ is well-defined; (ii) both positive and negative (or at least non-identical) signals are present, preventing the trivial zero-gradient case where all $\hat A_i$ are identical. Consequently, $\tilde g(\theta)$ is a non-degenerate direction whenever useful learning signal exists.

\paragraph{Asymptotic unbiasedness with vanishing threshold.}
If the threshold decays $\tau_{\text{adv}}^{(t)}\downarrow 0$ and the law of $\hat A_i$ has a continuous density at $0$,
then the discard probability $\mathbb{P}(|\hat A_i|\le \tau_{\text{adv}}^{(t)})\to 0$, and
\begin{equation}
\lim_{t\to\infty}\Big\|\ \mathbb{E}[\tilde g_{\,t}(\theta)]-\mathbb{E}[g(\theta)]\ \Big\| \;=\;0,
\end{equation}
i.e., the dynamic filtering becomes asymptotically unbiased while retaining finite-time variance-reduction benefits.

\paragraph{Summary.}
Dynamic filtering deletes rollouts whose contributions are provably negligible (zero or $O(\tau_{\text{adv}})$),
thereby reducing variance and compute without altering the expected update in the limit $\tau_{\text{adv}}\!\to\!0$; using the same $1/G$ normalization as \eqref{eq:rl-obj} avoids spurious scaling bias.

\subsubsection{Convergence Considerations for the Clipped Sequence-Level Objective}
Consider the surrogate objective $\mathcal{J}_{\text{RL}}(\theta)$ in \eqref{eq:rl-obj}, where rewards are bounded in $[0,1]$ and the sequence-level importance ratios are dual-clipped to $[\,1-\varepsilon_{\text{low}},\,1+\varepsilon_{\text{high}}\,]$.

\paragraph{Assumptions.}
\begin{itemize}
\item[(A1)] \textbf{Bounded scores.} There exists $B_\pi<\infty$ such that for all histories $(q,o_{<t})$ and tokens $o_t$,
$\big\|\nabla_\theta \log \pi_\theta(o_t\!\mid\!q,o_{<t})\big\|\le B_\pi$.
\item[(A2)] \textbf{Bounded rewards \& finite clipping.} For each rollout $o_i$, $R_i\in[0,1]$ and 
$r_{\mathrm{seq},i}(\theta)\in[\,1-\varepsilon_{\text{low}},\,1+\varepsilon_{\text{high}}\,]$ with $0<\varepsilon_{\text{low}},\varepsilon_{\text{high}}<\infty$.
\item[(A3)] \textbf{Non-degenerate batch dispersion.} On accepted groups, $\mathrm{Var}(\{R_i\}_{i=1}^G)\ge \tau_{\text{var}}>0$, so $\hat A_i = (R_i-\bar R)/\mathrm{std}(R)$ are well-defined.
\item[(A4)] \textbf{Vanishing filtering.} $\tau_{\text{adv}}^{(t)}\downarrow 0$ and the law of $\hat A_i$ has a continuous density at $0$, so $\mathbb{P}(|\hat A_i|\le \tau_{\text{adv}}^{(t)})\to 0$.
\item[(A5)] \textbf{Stepsizes.} Robbins--Monro conditions: $\sum_{t} \eta_t=\infty$ and $\sum_{t} \eta_t^2<\infty$.
\end{itemize}

\paragraph{Lemma 6 (Bounds on per-sample gradients and second moments).}
Let $o=(o_1,\dots,o_T)$ with $T:=|o|$, and let $h_t:=(q,o_{<t})$.
Let $r_{\mathrm{seq}}(\theta)$ denote the (clipped) sequence ratio. Then
\begin{equation}
\begin{aligned}
\nabla_\theta r_{\mathrm{seq}}(\theta)
&=
\frac{r_{\mathrm{seq}}(\theta)}{T}\sum_{t=1}^{T}\nabla_\theta \log \pi_\theta(o_t\mid h_t),\\
\big\|\nabla_\theta r_{\mathrm{seq}}(\theta)\big\|
&\le (1+\varepsilon_{\text{high}})\,B_\pi.
\end{aligned}
\end{equation}

Moreover, the PPO-style term is piecewise smooth and its gradient magnitude is bounded by $C_1:= (1+\varepsilon_{\text{high}})B_\pi\,|\hat A|$; together with (A3), $|\hat A|\le \frac{1}{\sqrt{\tau_{\text{var}}}}$ yields a uniform second-moment bound $\mathbb{E}\big[\|\nabla_\theta \ell_i(\theta)\|^2\big]\le C_2<\infty$.

\paragraph{Lemma 7 (Asymptotic unbiasedness under vanishing filtering).}
Let $g(\theta)$ denote the full (unfiltered) stochastic gradient and
$\tilde g_{\tau}(\theta)=\frac{1}{G}\sum_{i:\,|\hat A_i|>\tau} g_i(\theta)$ the filtered version with $\frac{1}{G}$ normalization.
Under (A4) and the bounded second moments above,
\begin{equation}
\lim_{\tau\downarrow 0}\ \big\|\,\mathbb{E}[\tilde g_{\tau}(\theta)]-\mathbb{E}[g(\theta)]\,\big\| \;=\;0
\quad \text{for all }\theta.
\end{equation}

\paragraph{Theorem 1 (Convergence to a stationary point of the surrogate).}
Suppose (A1)–(A5) hold. Then the iterates of stochastic gradient ascent on $\mathcal{J}_{\text{RL}}(\theta)$ with the dynamic filtering scheme converge almost surely to the set of stationary points of the surrogate objective.

\emph{Proof sketch.}
By Lemma~6 and the reward boundedness (Lemma~3), the stochastic gradients have uniformly bounded second moments; the objective is bounded and piecewise smooth (kinks of measure zero). Lemma~7 guarantees that the bias due to filtering vanishes as $\tau_{\text{adv}}^{(t)}\to 0$. Therefore the noisy gradient process forms a Robbins--Monro stochastic approximation with asymptotically unbiased gradients and square-summable noise, yielding a.s.\ convergence to stationary points of $\mathcal{J}_{\text{RL}}$ (e.g., Kushner--Yin/Bottou).

\paragraph{Remarks.}
(i) The min-with-clipping introduces bias w.r.t.\ the \emph{true} off-policy objective, but ensures variance control and stability; the theorem concerns the surrogate we optimize. (ii) Sequence-level ratios and sequence-level clipping align the gradient scale with the sequence reward, avoiding token/sequence granularity mismatch and contributing to the boundedness needed above. (iii) In practice, we keep $\tau_{\text{var}}$ and the clip window fixed and decay $\tau_{\text{adv}}$, which satisfies the lemmas’ conditions and matches our training protocol.

\end{document}